\documentclass[lettersize,journal]{IEEEtran}

\usepackage{amsmath,amssymb,amsfonts,amsthm}
\usepackage{graphicx}
\usepackage{cite}
\usepackage{textcomp}
\usepackage{subcaption}
\usepackage{booktabs,tabularx,array,multirow}
\usepackage{xurl}
\usepackage{leftindex}
\usepackage{wrapfig}
\usepackage{xspace}
\usepackage{placeins}
\usepackage{xcolor}
\usepackage{enumitem}
\usepackage[hidelinks]{hyperref}
\usepackage{algorithm,algpseudocode}
\usepackage[font=footnotesize]{caption}

\newcommand{\SE}{SE(3)}
\newcommand{\SO}{SO(3)}
\newcommand{\fg}{f_{_G}}
\newcommand{\eg}{e_{_G}}
\newcommand{\calO}{\mathcal{O}}
\newcommand{\gedf}{g_{_{\mathrm{EDF}}}}
\newcommand{\Ad}{\text{Ad}}
\newcommand{\EDF}{f_{\varphi}}

\theoremstyle{definition}
\newtheorem{lemma}{Lemma}

\newtheorem{proposition}{Proposition}

\newtheorem{corollary}{Corollary}

\title{EquiContact: A Hierarchical SE(3) Vision-to-Force Equivariant Framework for Spatially Generalizable Contact-Rich Tasks}
\author{Joohwan Seo, Arvind Kruthiventy, Soomi Lee, Megan Teng, Seoyeon Choi, Xiang Zhang, \\
Jongeun Choi, and Roberto Horowitz%
\thanks{J. Seo, A. Kruthiventy, S. Lee, M. Teng, S. Choi, X. Zhang, and R. Horowitz are with the University of California, Berkeley. J. Choi is with Yonsei University.}}

\begin{document}
\maketitle

\begin{abstract}
This paper presents EquiContact, a hierarchical $\SE$ vision-to-force equivariant framework for spatially generalizable contact-rich manipulation.
EquiContact consists of a high-level equivariant vision module that estimates global task reference frames and a localized compliant policy (LCP) anchored on these frames for fine contact-rich execution.
The LCP uses wrist-mounted RGB observations, a geometrically consistent error vector (GCEV), and end-effector-frame force--torque feedback to generate relative end-effector pose commands and admittance gains, which are executed through geometric admittance control (GAC).
We show that the complete EquiContact pipeline is $\SE$ equivariant under equivariant reference-frame estimation and left-invariant visual features, without requiring the LCP itself to be an explicitly equivariant neural network.
To address the visual-invariance condition in practice, we introduce object-centric visual preprocessing and systematically investigate the learned visual representation.
With the LCP trained on demonstrations from a single task configuration, the full pipeline succeeds in $39/40$ circular-peg and $40/40$ triple-peg insertion trials with $1\;\mathrm{mm}$ clearance on a real robot, across translated-flat, $30^\circ$ tilted, and $45^\circ$ tilted configurations.
With ground-truth reference frames, the LCP also succeeds at triple-peg tilts of $90^\circ$ and $135^\circ$ and on a surface-wiping task.
\end{abstract}

\begin{IEEEkeywords}
Contact-rich task, Imitation Learning, Compliant Control, Geometric Control,
SE(3)-Equivariance
\end{IEEEkeywords}

\begin{figure*}[!t]
    \centering
    \includegraphics[width=0.97\textwidth]{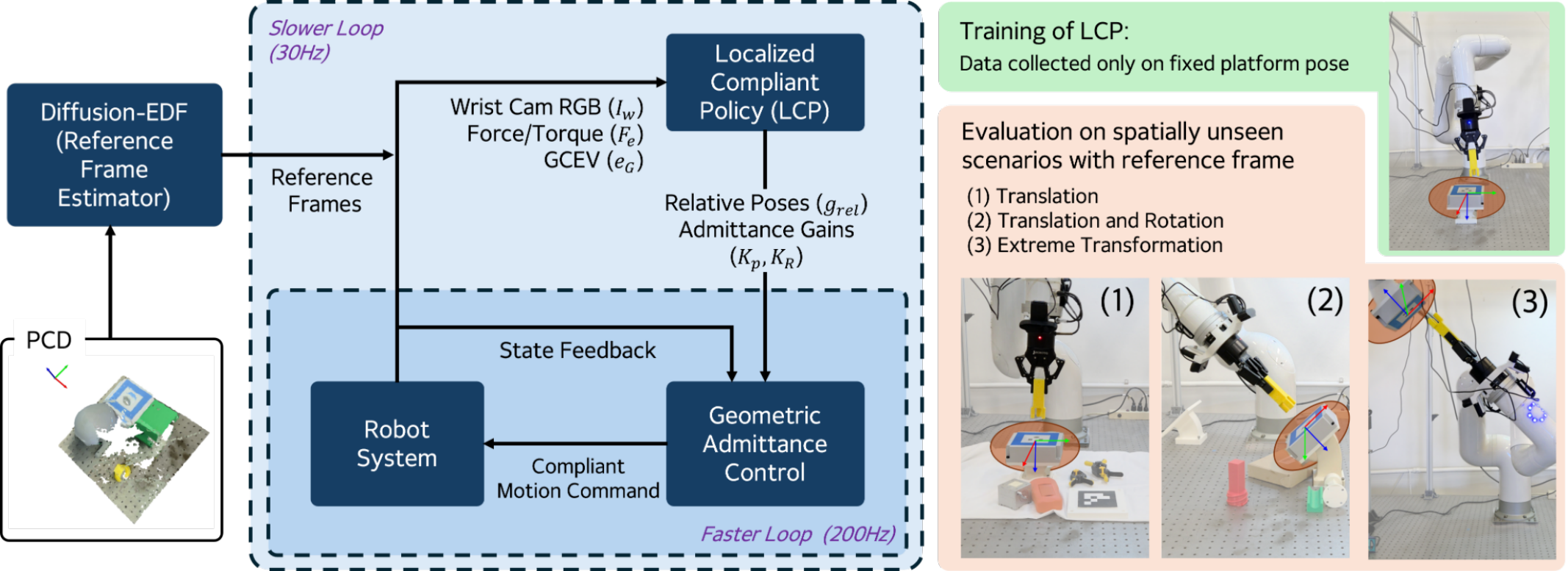}
    \vspace{-0.2em}
    \caption{Overview of EquiContact. \textbf{(Left)} The high-level Diff-EDF module estimates task reference frames from the scene point cloud. A localized compliant policy (LCP) is anchored on these frames and uses GCEV, body-frame force--torque feedback, and object-centric wrist RGB observations to generate relative pose commands and admittance gains, which are executed by GAC. \textbf{(Right)} The LCP is trained from demonstrations collected at a fixed task configuration and is evaluated at $\SE$-transformed task configurations with background variations through the estimated task reference frame.}
    \label{fig:equi_contact}
    \vspace{-3pt}
\end{figure*}

\section{Introduction}\label{sec:introduction}
Imitation learning has recently shown significant success in expanding the capabilities of real-world robotic manipulation~\cite{o2024open,black2024pi_0,zeng2020transporter,shridhar2022cliport,chi2023diffusion,zhao2023learning,shafiullah2022behavior}.
Similar to trends observed in large language models, there is growing evidence that large-scale datasets can enable generalizable vision-based robotic policies~\cite{kim2024openvla,o2024open}.
However, generalization across different task positions and orientations often relies on spatially diverse training demonstrations.
An alternative approach is to exploit geometric symmetry, particularly equivariance, to achieve spatial generalization without requiring demonstrations across every task configuration~\cite{seo2025se,seo2023contact}.


Contact-rich tasks such as precise assembly require maintaining stable contact without applying excessive force. A common approach is to execute a reference motion with an impedance controller that governs how the robot responds to contact forces~\cite{hogan1985impedance}. A learned policy can use sensory feedback to refine the commanded motion and adjust impedance gains during execution~\cite{kamijo2024learning,kim2026compvla}. 
For a policy to be spatially transferable on these contact-rich tasks, both its commanded motion and its contact response should transform with the task.
One approach is to collect or generate demonstrations across task configurations~\cite{mandlekar2023mimicgen}, but these demonstrations need  also to capture the contact interactions at those configurations. We instead exploit $\SE$ symmetry to transfer a policy learned at a single nominal configuration while preserving its task-relative contact behavior at unseen configurations.

In this work, we propose \emph{EquiContact}, a hierarchical $\SE$ vision-to-force equivariant framework for spatially generalizable contact-rich manipulation. 
The first stage of EquiContact is a high-level $\SE$-equivariant vision module, implemented using Diffusion Equivariant Descriptor Fields (Diff-EDF)~\cite{diff_edf}, which estimates \emph{task reference frames} from point-cloud observations defined on the global frame.
These estimated frames provide global reference information for the second-stage policy.

The second-stage policy performs fine end-effector control and manages contact interactions using local sensory inputs: force--torque (FT) measurements expressed in the end-effector frame, the $\SE$ left-invariant Geometrically Consistent Error Vector (GCEV)~\cite{seo2023contact} between the current end-effector frame and the estimated task reference frame, and wrist-camera images.
The policy outputs relative end-effector pose commands and admittance gains, which are executed through $\SE$ left-invariant Geometric Admittance Control (GAC)~\cite{seo2023geometric}.
We refer to this observations and actions parameterization defined relative to the current end-effector frame as a local representation,
and therefore we call this second-stage policy the \emph{localized compliant policy} (LCP).
Before the LCP is activated, the end-effector is positioned near the globally estimated task reference frame. During execution, the LCP continuously uses local sensory inputs to perform local refinement.
We refer to this combination of coarse positioning and reference-relative feedback as \emph{anchoring} the localized policy on the globally estimated reference frame.

Prior work on contact-rich manipulation showed that a policy defined in the end-effector frame is $\SE$-equivariant in the spatial frame if it is invariant under a left transformation jointly applied to the task and the end-effector~\cite{seo2023contact}.
We extend this result to a vision-to-force pipeline in which the task reference frame is estimated from vision and the LCP uses wrist-camera images to handle contact interactions.
A sufficient condition for the LCP to remain invariant under a left $\SE$ transformation is that the visual features produced by the wrist-camera encoder remain invariant under the same transformation.
However, this property is not guaranteed by the wrist-camera observations, which may change due to different backgrounds, illumination, and surrounding scene content, even when the relative task geometry remains unchanged.
In the proposed framework, we identify left-invariance of the visual representation as a condition for pipeline-level equivariance and introduce object-centric visual preprocessing to approximate this property.
We further investigate the practical realization of the visual representation through a systematic design study.



Under equivariant reference-frame estimation and left-invariant visual features, the LCP produces the same relative pose and compliance commands when the task and end-effector undergo a common left transformation.
Combined with GAC, the resulting elastic wrench that governs contact interaction transforms equivariantly when expressed in the spatial frame.
Thus, the low-level LCP inherits the equivariance of the high-level reference-frame estimator, making the overall vision-to-force pipeline $\SE$ equivariant under these conditions.
We refer to this pipeline-level property as \emph{induced equivariance}, since the equivariance of the complete framework is obtained through the geometric structure of its components without requiring the low-level neural policy itself to be explicitly equivariant.

An overview of EquiContact is shown in Fig.~\ref{fig:equi_contact}.
We emphasize that our primary contribution lies in the overall structural framework rather than in the specific choice of algorithms.
The high-level reference-frame estimator and the low-level visuomotor policy can be implemented using different learning architectures, provided that their geometric transformation properties remain compatible.

The main contributions of this paper are as follows:
\begin{enumerate}[leftmargin=*]
\setlength{\itemsep}{2pt}

\item We propose EquiContact, which anchors a localized compliant policy to an equivariantly estimated task reference frame. We show that, under equivariant reference-frame estimation and left-invariant visual features, the elastic wrench is $\SE$-equivariant in the spatial frame without requiring an explicitly equivariant low-level network.

\item We identify wrist-camera features as a source of left-invariance error and use object-centric preprocessing to reduce background variation. A sequential real-robot design study shows that encoder adaptation and background painting improve transfer to tilted task configurations.

\item We evaluate EquiContact on real-robot peg-in-hole tasks with $1\,\mathrm{mm}$ clearance, using LCP demonstrations collected at a single fixed configuration for each task.
The full pipeline succeeds in $39/40$ circular-peg trials and $40/40$ triple-peg trials across translated-flat, $30^\circ$ tilted, and $45^\circ$ tilted configurations.
With ground-truth task reference frames, the LCP also transfers to triple-peg tilts of $90^\circ$ and $135^\circ$ and to surface wiping.
\end{enumerate}

The remainder of this paper is organized as follows. Section~\ref{sec:related_works} reviews related work. Section~\ref{sec:problem_formulation} formulates the problem, and Section~\ref{sec:background} introduces the necessary mathematical preliminaries. 
Section~\ref{sec:solution_approach} presents the proposed EquiContact framework
and establishes its vision-to-force equivariance properties. Section~\ref{sec:lcp_design} investigates the empirical design of the
localized compliant policy. Section~\ref{sec:experimental_result} presents the real-robot experimental results, and Section~\ref{sec:conclusion} concludes the paper. Additional implementation details and experimental results are provided in the Appendix.

\section{Related Work}\label{sec:related_works}
\textbf{Visuomotor Policy Learning.}
Generative models have become widely used for learning visuomotor policies from demonstrations. Representative approaches include Action Chunking with Transformers (ACT)~\cite{zhao2023learning}, Diffusion Policy~\cite{chi2023diffusion}, and conditional flow-matching policies~\cite{black2024pi_0,ding2025rfmp}. These approaches employ different generative architectures to model visuomotor policies.
EquiContact adopts a conditional Flow Matching action-generation model following the GR00T N1 architecture \cite{gr00tn1_2025}, while focusing on the geometric representation of observations and actions for spatial generalization.

\textbf{Force-Aware and Compliant Visuomotor Policies.}
Recent work has incorporated force feedback and compliant control into learned manipulation policies. CompACT~\cite{kamijo2024learning} learns variable compliance through action chunking, while Adaptive Compliance Policy (ACP)~\cite{hou2025adaptive} learns task-dependent compliance for diffusion-based manipulation. 
CompVLA~\cite{kim2026compvla} extends vision-language-action
models to jointly predict motion and compliance commands,
which are executed through geometric impedance control (GIC)
with a Lie-algebra-based potential function~\cite{seo2024comparison}.
Other approaches incorporate force sensing into policy learning,
including \cite{he2025foar, yu2025forcevla, choi2026umift}.
These studies demonstrate the importance of force-aware policies and variable compliance for contact-rich manipulation. Complementing these developments, EquiContact investigates how a learned compliant policy can be structured to preserve $\SE$ transformation properties across global perception, local policy generation, and force-level execution.

\textbf{Equivariant Methods.}
Earlier approaches incorporated $SO(2)$ or $\SE$ symmetries into policy architectures for pick-and-place tasks using images or point clouds~\cite{ndf,diff_edf}. EquAct~\cite{zhu2025equact} extends this direction through equivariant keyframe prediction with language conditioning. 
Other methods formulate equivariant denoising processes for pose-trajectory generation~\cite{yang2024equibot,tie2024seed,wang2024equivariant,action_flow}.
Recent approaches also explore 3D equivariance from 2D image observations~\cite{hu2025equivariant,klee2026raven}. 
Complementing these approaches that impose equivariance directly through the neural network's property, EquiContact investigates how equivariance can be preserved through hierarchical policy composition and extended to force-level execution.

\textbf{Reference-Frame Representations.}
The choice of reference frame influences the spatial generalization of learned manipulation policies. Several approaches express policy observations or actions relative to object or task frames~\cite{chentool,rana2025learning,zhao2025hierarchical}, while wrist-centric representations have also been explored for symmetry-aware policy learning~\cite{wang2025practical}.
EquiContact uses an equivariantly estimated task reference to anchor a localized policy whose geometric observations and actions are expressed in the end-effector frame.

\textbf{Object-Centric Representations.}
Recent work has explored object-centric visual representations to improve visuomotor policy generalization. Approaches include guiding policy attention toward task-relevant regions~\cite{sun2026artificial} and grouping visual features into object-like slots~\cite{chapin2026spotlighting}. Segmentation-based approaches explicitly construct policy inputs from masked RGB images~\cite{shi2024composing} or semantically repainted observations~\cite{ding2026coordination}. These studies motivate restricting visual representations to task-relevant information. EquiContact adopts foreground masking and connects object-centric visual invariance to the conditions required for $\SE$-equivariant policy composition.

\section{Problem Formulation} \label{sec:problem_formulation}
%

\begin{figure}[t]
    \centering
    \includegraphics[width=0.85\linewidth]{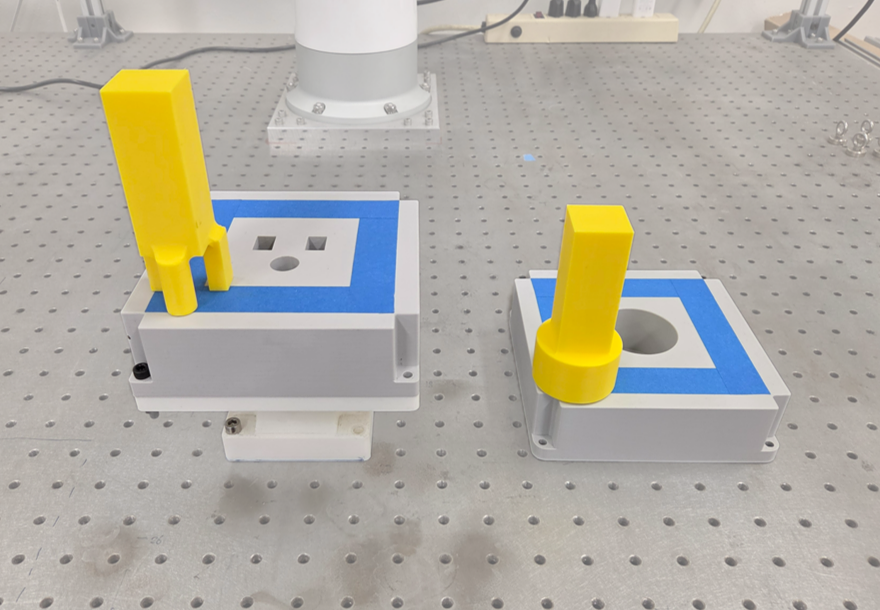}
    \caption{Triple (Left) and circular (Right) pegs and corresponding platforms with hole. Both tasks have $1\;\mathrm{mm}$ clearance.}
    \label{fig:circular_and_triple_pegs}
\end{figure}

We study spatial generalization of a policy for contact-rich manipulation that is trained on demonstrations collected at a single nominal task configuration.
Our primary focus is peg-in-hole assembly. The robot first grasps the peg, then aligns it with the hole and regulates contact forces during insertion. We assume that the peg is upright in the scene, although its position can change across trials.
We consider circular and orientation-sensitive triple-peg assemblies, both with $1\,\mathrm{mm}$ clearance (Fig.~\ref{fig:circular_and_triple_pegs}).

Let $g_{\mathrm{ref}}\in\SE$ denote the true task reference frame at the nominal configuration. At test time, a rigid transformation $g_l\in\SE$ changes the task reference frame to $g_l g_{\mathrm{ref}}$. For insertion, this corresponds to translating or tilting the platform. The objective is to reuse the policy learned at the nominal configuration under these transformations without collecting additional demonstrations at each test configuration.

\begin{figure}[t]
    \centering
    \includegraphics[width=0.95\linewidth]{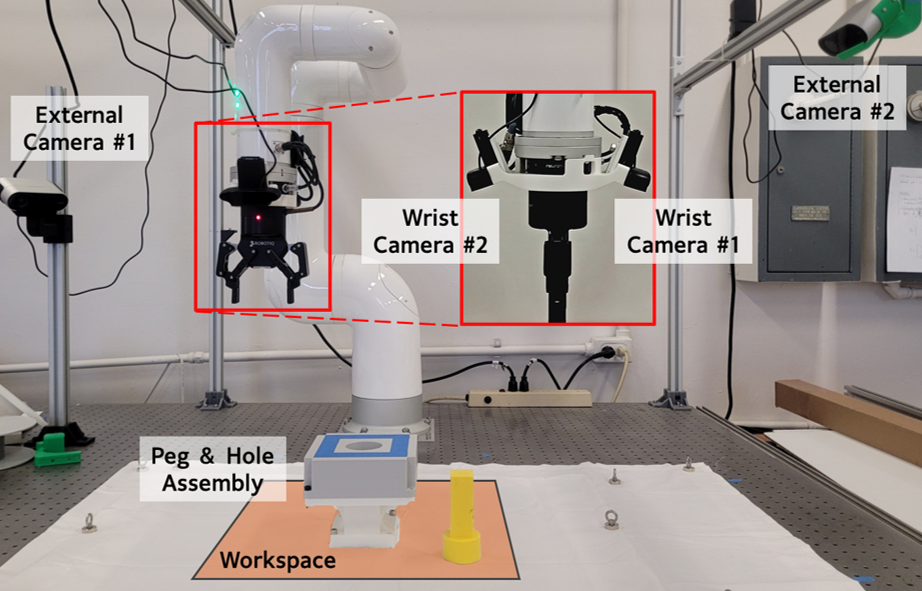}
    \vspace{-2pt}
    \caption{
    Experimental peg-in-hole setup. Two external cameras provide global point-cloud observations, while two wrist-mounted cameras provide local RGB feedback.
    }
    \label{fig:overall_scene}
\end{figure}

\begin{figure}[t]
    \centering
    \includegraphics[width=\columnwidth]{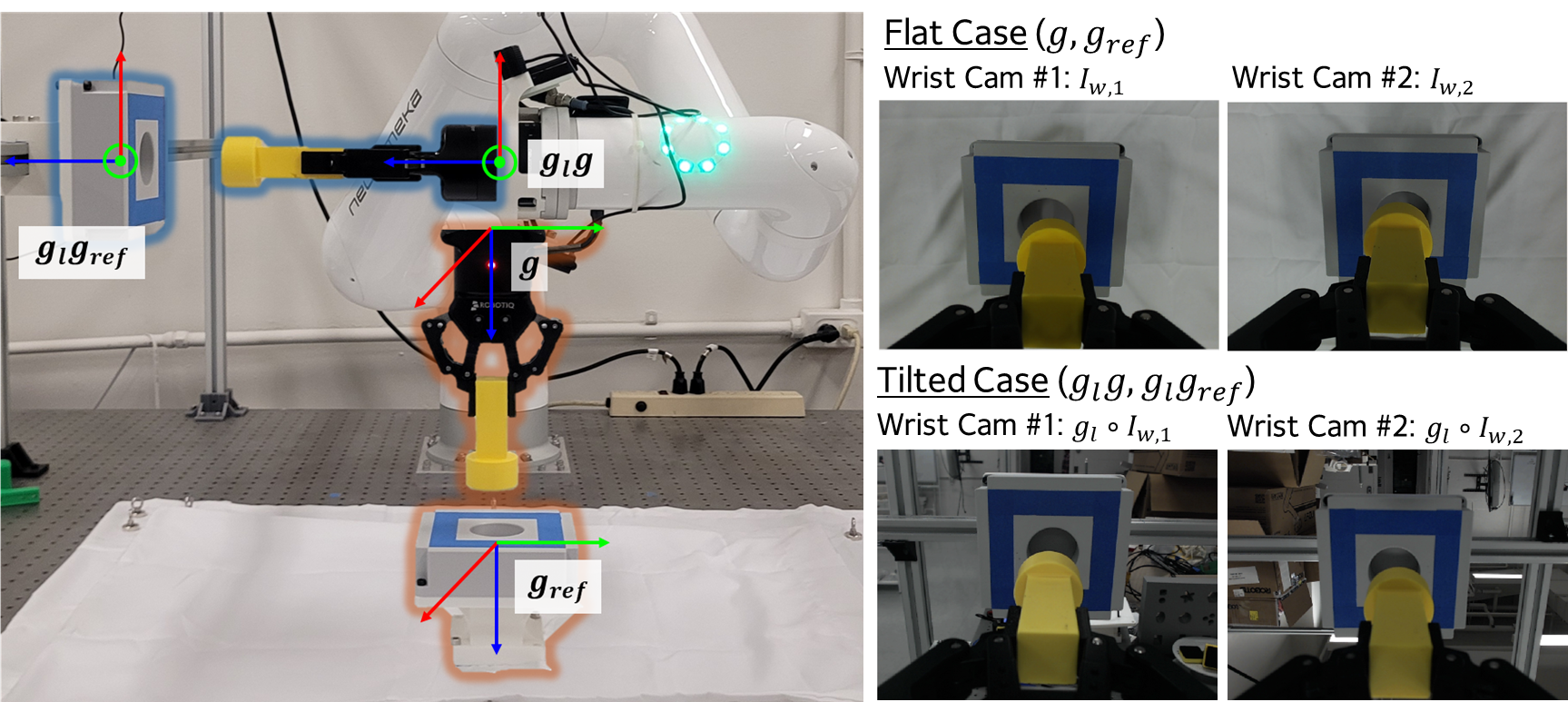}
    \caption{Wrist-camera observations under a task transformation.
    The end-effector and task frame undergo the same rigid transformation $g_l$,
    preserving their relative pose. The surrounding scene remains fixed,
    so the wrist camera can see a different background.}
    \label{fig:group_action_on_image}
\end{figure}

The true task reference frame $g_{\mathrm{ref}}$ is unknown at test time. The external cameras observe the task object and its surroundings in the form of point cloud, and a global estimator uses the resulting point cloud to obtain a coarse estimate of the task frame (Fig.~\ref{fig:overall_scene}). An error in the estimated task reference frame can leave the peg misaligned with the hole. The robot therefore uses local feedback to correct the alignment during execution. Let $g(k)\in\SE$ denote the end-effector pose at sampling instant $k$. Even when the end-effector moves to $g_l g(k)$ and preserves its pose relative to the transformed task, the wrist cameras can see a different background because the surrounding scene remain fixed, as shown in Fig.~\ref{fig:group_action_on_image}.

At each sampling instant, the policy receives a local observation $o(k)$ comprising geometric feedback relative to the estimated task frame, wrist images, and end-effector-frame force/torque feedback, which will be formally defined later in the paper. The policy $\pi_\theta$ maps this observation to an action $a(k)=\pi_\theta(o(k))$. The action specifies a relative end-effector motion command and stiffness gains for compliant execution.

\section{Background}\label{sec:background}
\subsection{$\SE$ Left Invariance and Equivariance} \label{sec:left_invariance}
We first introduce the notions of left invariance and equivariance used throughout the paper. 
Let $f:\mathcal{X}\rightarrow\mathcal{Y}$ be a function where $\mathcal{X}$ and $\mathcal{Y}$ denote the domain and codomain, respectively, and let $g_l\in\SE$ act on domain or codomain through the group action $\circ$. 
The function $f$ is \emph{left-invariant} if
\begin{equation} \label{eq:left_invariance}
    f(g_l\circ x)=f(x), \qquad \forall g_l\in\SE,
\end{equation}
and is \emph{left-equivariant} if
\begin{equation} \label{eq:left_equivariance}
    f(g_l\circ x)=g_l\circ f(x), \qquad \forall g_l\in\SE,
\end{equation}
where the group action on the input and outputs is determined by the representation of $\mathcal{X}$ and $\mathcal{Y}$~\cite{seo2025se}. 
For an end-effector pose $g\in\SE$, the left action is a standard matrix multiplication:
\begin{equation}
    g_l\circ g = g_l \cdot g.
\end{equation}
Thus, left invariance describes quantities that remain unchanged when the spatial configuration is transformed by $g_l$, whereas left equivariance describes quantities that transform consistently with $g_l$. 
The EquiContact formulation combines both properties: locally defined geometric and force quantities are represented to be left-invariant, while the high-level task reference frame transforms equivariantly with the task.

\subsection{Force-Level Execution via Geometric Admittance Control (GAC)} \label{sec:GAC}
Geometric admittance control (GAC) \cite{seo2023geometric, seo2024comparison} converts locally predicted pose and gain commands into compliant body-frame force interactions while preserving the geometric structure used by the policy.
Given the current end-effector frame $g = (p,R)$ and the desired end-effector frame $g_d = (p_d, R_d)$, provided by the low-level policy \eqref{eq:LCP}, the desired end-effector dynamics in the GAC formulation are given by
\begin{equation} \label{eq:desired_dynamics}
    M \dot{V}^b + K_d V^b + \fg = F_e, 
\end{equation}
where
\begin{equation} \label{eq:GIC_wrench}
    \fg = \begin{bmatrix}
    f_p \\ f_R
    \end{bmatrix} =\begin{bmatrix}
    R^\top R_d K_p R_d^\top (p - p_d)\\
    (K_R R_d^\top R - R^\top R_d K_R)^\vee
    \end{bmatrix},
\end{equation}
where $M \in \mathbb{R}^{6\times6}$ is a symmetric positive definite desired inertia matrix, $K_d \in \mathbb{R}^{6\times6}$ is a symmetric positive definite damping matrix, $F_e\in \mathbb{R}^6$ is the FT sensor signal in the end-effector body frame and $V^b \in \mathbb{R}^6$ is a body-frame end-effector velocity. $(\cdot)^\vee$ is a vee map that maps skew-symmetric matrix to $\mathbb{R}^3$ vector . Further, $\fg = \fg(g,g_d, K_p, K_R) \in \mathbb{R}^6$ is an elastic wrench, where $K_p, K_R \in \mathbb{R}^{3\times3}$ are symmetric positive definite learned stiffness matrices for the translational and rotational dynamics, respectively. A key property of GAC is that for any arbitrary transformation $g_l \in \SE$, $ \fg(g_l g,g_lg_d, K_p, K_R)   = \fg(g,g_d, K_p, K_R)$ (see Lemma~\ref{lemma:1}). Thus, the elastic wrench is left-invariant under a common left transformation of the current and desired end-effector frames. 
Refer to Appendix~\ref{app:GAC_implementation} for the detailed implementation of GAC.

\subsection{Equivariant Reference-Frame Estimation (Diff-EDF)} \label{sec:Diff-EDF}
Diffusion-Equivariant Descriptor Field (Diff-EDF)~\cite{diff_edf} is an $\SE$-equivariant reference-frame estimator for pick-and-place tasks. In EquiContact, Diff-EDF serves as a high-level vision module that provides a coarse estimate of the target reference frame $\gedf$ for the downstream localized policy.
Given a scene point cloud $\calO^{\mathrm{scene}}$ and a gripper point cloud $\calO^{\mathrm{grasp}}$, Diff-EDF outputs an estimated target pose $\gedf \in \SE$:
\begin{equation} \label{eq:EDF}
    \gedf = \EDF(\calO^{\mathrm{scene}}, \calO^{\mathrm{grasp}}).
\end{equation}

The equivariant architecture of Diff-EDF, acts on the entire scene point cloud. For a rigid transformation $g_l\in\SE$, its \emph{guaranteed} scene-level transformation property is
\begin{equation} \label{eq:EDF_scene_equivariance}
    \EDF(g_l\circ\calO^{\mathrm{scene}},\calO^{\mathrm{grasp}})
    =g_l \cdot \EDF(\calO^{\mathrm{scene}},\calO^{\mathrm{grasp}}),
\end{equation}
where $g_l\circ\calO$ applies the transformation to every point in $\calO$.

In our setting, the task object moves while other parts of the scene remain fixed. Let $\calO^{\mathrm{obj}}\subset\calO^{\mathrm{scene}}$ contain the task object and let
$\calO^{\mathrm{others}}=\calO^{\mathrm{scene}}\setminus\calO^{\mathrm{obj}}$.
Writing the estimator input as
$\EDF(\calO^{\mathrm{obj}},\calO^{\mathrm{grasp}},\calO^{\mathrm{others}})$,
the \emph{desired} property of Diff-EDF can be written by:
\begin{equation} \label{eq:equivariance_diff_edf}
    \EDF(g_l\circ\calO^{\mathrm{obj}},
         \calO^{\mathrm{grasp}},\calO^{\mathrm{others}})
    =g_l\cdot \EDF(\calO^{\mathrm{obj}},
              \calO^{\mathrm{grasp}},\calO^{\mathrm{others}}).
\end{equation}
Scene-level equivariance does not guarantee~\eqref{eq:equivariance_diff_edf}, since \(\calO^{\mathrm{others}}\) remains fixed. In practice, Diff-EDF can approximately follow the task object when its prediction depends primarily on \(\calO^{\mathrm{obj}}\). We therefore use~\eqref{eq:equivariance_diff_edf} as a condition in the subsequent analysis of pipeline-level equivariance.


\section{EquiContact}\label{sec:solution_approach}
\subsection{EquiContact Overview}
The core insight of EquiContact is to use an equivariant high-level vision module to determine task reference frames and to anchor a localized low-level policy on these frames for fine-grained contact-rich execution.
In our implementation, Diff-EDF estimates task reference frames
$\gedf^t=(p^t_{_{\mathrm{EDF}}},R^t_{_{\mathrm{EDF}}})$, $t\in\{G,I\}$,
from scene point clouds, where $G$ and $I$ denote grasping and insertion, respectively.
For each stage, the manipulator first moves the end-effector near $\gedf^t$.
The remaining contact-rich execution is then handled by a \emph{Localized Compliant Policy (LCP)} using wrist-mounted RGB observations, end-effector-frame force--torque (FT) measurements, and the geometrically consistent error vector (GCEV) between the current end-effector frame and $\gedf^t$.
The LCP outputs relative end-effector pose commands and geometric admittance gains, which are executed through geometric admittance control (GAC).

From the perspective of the LCP, a left transformation of both the task and the end-effector should not change the local execution problem when their relative configuration is preserved.
Accordingly, identical local observations should produce identical relative motion and compliance commands.
We refer to this property as $\SE$ left-invariance of the localized policy, following the definition in Sec.~\ref{sec:left_invariance}.
When the reference-frame estimator is $\SE$ equivariant, this left-invariant low-level policy can be anchored at the transformed task frame, yielding the vision-to-force equivariance stated in Sec.~\ref{sec:equivariance}.
Unless otherwise stated, we write $\gedf$ for either a grasping or an insertion task reference frame.

\subsection{Localized Compliant Policy (LCP)} \label{Sec:LCP}
The localized compliant policy (LCP) is the low-level feedback policy in EquiContact. 
Its role is to refine the coarse task reference frame using local visual, proprioceptive, and force feedback and to produce commands for compliant contact execution.
The geometric input and output representations follow the design principles in~\cite{seo2023contact}: quantities used for low-level feedback, i.e., observations, are defined relative to the current end-effector frame, and the resulting pose command is also expressed relative to that frame.

Write $g(k)=(p(k),R(k))$, where $p(k)\in\mathbb{R}^3$ and $R(k)\in\SO$ are the position and orientation of the end-effector.
Let $\gedf=(p_{_{\mathrm{EDF}}},R_{_{\mathrm{EDF}}})\in\SE$ denote the task reference frame estimated by Diff-EDF.
The GCEV between these two frames is
\begin{equation} \label{eq:GCEV}
    \eg(g,\gedf)=
    \begin{bmatrix}
        R^\top(p-p_{_{\mathrm{EDF}}})\\
        (R_{_{\mathrm{EDF}}}^\top R-R^\top R_{_{\mathrm{EDF}}})^\vee
    \end{bmatrix},
\end{equation}
where $(\cdot)^\vee$ denotes the vee-map from a skew-symmetric matrix to $\mathbb{R}^3$~\cite{murray1994mathematical}.
The vector $\eg\in\mathbb{R}^6$ is left-invariant, i.e., $\eg(g_lg,g_l\gedf)=\eg(g,\gedf)$ for any $g_l\in\SE$~\cite{seo2023contact}.
The FT measurement $F_e(k)\in\mathbb{R}^6$ is expressed in the current end-effector frame and is therefore unchanged by a change of the spatial frame. The GCEV is one left-invariant representation of the relative pose $g^{-1}\gedf$; its translation paired with a six-dimensional rotation representation (\texttt{rot6d}) could also be used.

Let $I_w(k)$ denote the wrist-camera RGB observation. Notice that we denote $I_w$ as the concatenation of front and back wrist camera, without loss of generality.
The LCP observation is
\begin{equation}
    o(k)=\big(\eg(g(k),\gedf),F_e(k),I_w(k)\big).
\end{equation}
Given $o(k)$, the LCP predicts a relative desired pose $g_{\mathrm{rel}}(k)\in\SE$ together with the translational and rotational stiffness matrices $K_p(k)$ and $K_R(k)$ used by GAC, i.e., the output action is $a(k) = (g_{\mathrm{rel}},K_p,K_R)(k)$.
The desired pose in the spatial frame is obtained by composing the relative command with the current end-effector frame,
\begin{equation} \label{eq:LCP}
    (g_{\mathrm{rel}},K_p,K_R)(k)=\pi_\theta(o(k)),
    \quad
    g_d(k)=g(k)g_{\mathrm{rel}}(k).
\end{equation}
The tuple $(g_d,K_p,K_R)(k)$ is then supplied to GAC as described in Sec.~\ref{sec:GAC}. We will drop the time-instance $(k)$ for the rest of the paper unless it is specifically required.

In our implementation, $\pi_\theta$ uses a conditional flow-matching action-generation model following the GR00T N1 architecture \cite{gr00tn1_2025}.
The model generates an action chunk of horizon $H$, in the form of $(g_{\mathrm{rel}},K_p,K_R)_{i=1}^H$, which are then processed using temporal aggregation \cite{zhao2023learning} to obtain single-step command supplied to GAC.
The complete LCP, including object-centric preprocessing and DINOv3 feature extraction, updates at $30\,\mathrm{Hz}$.


\subsection{Object-Centric Visual Representation} \label{sec:vision_encoder_left_invariance}
As mentioned earlier, the GCEV $\eg$ and the body-frame FT measurement $F_e$ are left-invariant by construction, whereas the wrist-camera observation requires additional consideration.
Let $I_w(g,g_{\mathrm{ref}})$ denote the wrist-camera image observed at the end-effector pose $g$ and task reference pose $g_{\mathrm{ref}}$.
For a left transformation $g_l\in\SE$, we define the induced action on the wrist-camera observation as
\begin{equation} \label{eq:image_left_action}
    g_l\circ I_w(g,g_{\mathrm{ref}})
    := I_w(g_lg,g_lg_{\mathrm{ref}}).
\end{equation}
Figure~\ref{fig:group_action_on_image} shows how the wrist-camera image can change with the background even when the task object and the camera retain the same relative pose.
\begin{figure}
    \centering
    \includegraphics[width=\linewidth]{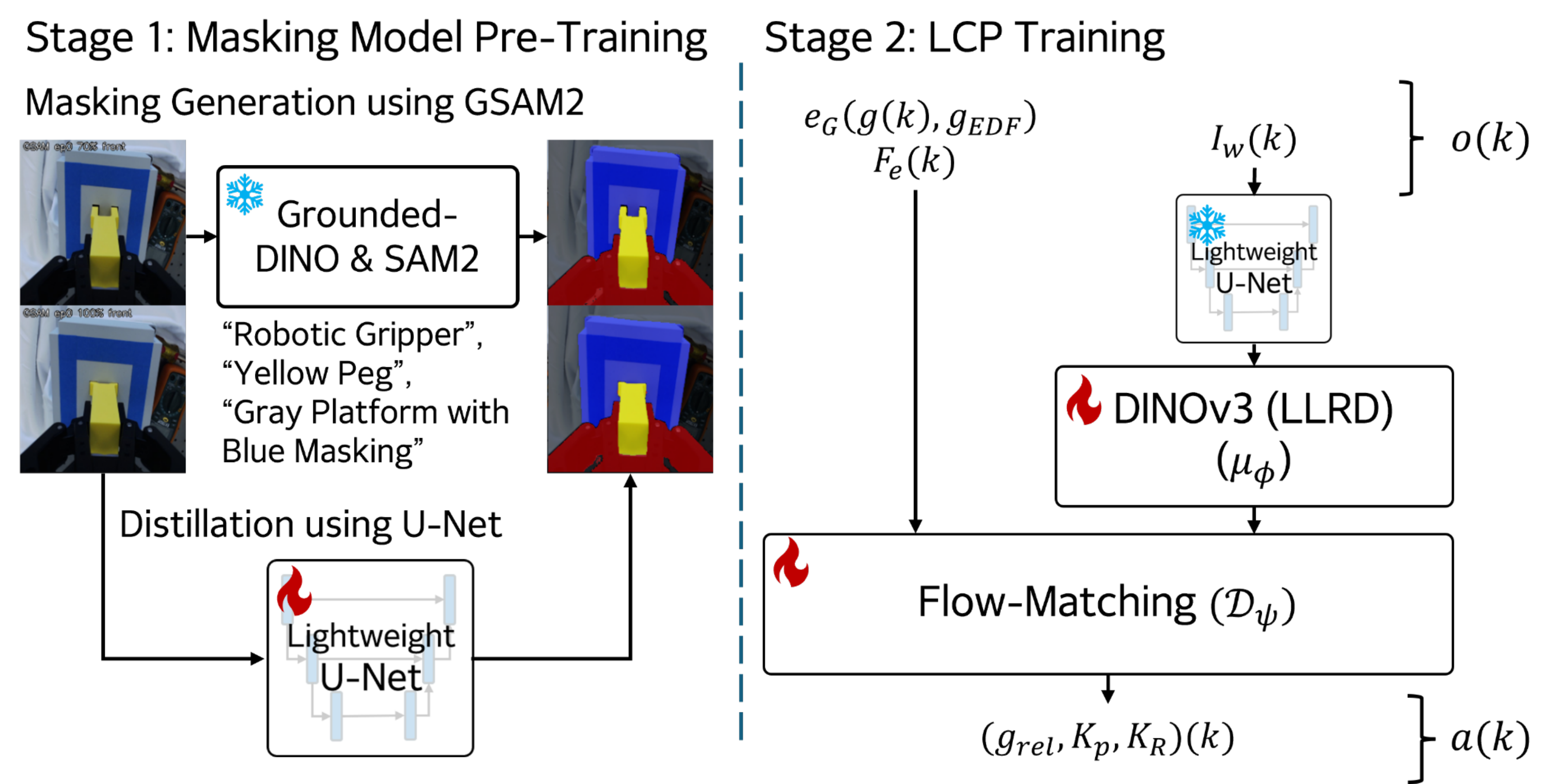}
    \caption{
    Two-stage LCP training pipeline. First, a lightweight U-Net is trained using generated segmentation labels (Grounded-DINO and SAM2 for peg insertion, and rule-based labels for wiping). Second, the LCP is trained through imitation learning with the U-Net frozen.
    }
    \label{fig:LCP_training}
\end{figure}

We therefore construct an object-centric wrist observation preprocessor that suppresses visual content unrelated to the local task geometry.
Let $\mathcal{P}$ denote the object-centric preprocessing operation and define 
\begin{equation}
    I_w^{\mathrm{obj}}=\mathcal{P}(I_w),
\end{equation}
where $I_w^{\mathrm{obj}}$ denotes the wrist camera image containing only task-related objects.
For each wrist-camera image, a lightweight U-Net predicts a segmentation mask.
For peg insertion, the U-Net is trained using labels generated offline by Grounded-DINO~\cite{liu2024groundingdino} and SAM2~\cite{ravi2025sam2}, which are too slow for the LCP's $30\;\mathrm{Hz}$ update rate. Wiping labels instead use color- and template-based rules. Only the lightweight U-Net is deployed; training details are given in Appendix~\ref{app:unet_training}.

The U-Net assigns each pixel to a class, such as peg, platform, robot gripper, or background.
The predicted multi-class segmentation is converted to a foreground mask $M_w(k)$ by assigning $1$ to pixels in the task-relevant classes and $0$ to the remaining pixels.
The preprocessing operation is given by:
\begin{equation} \label{eq:object_centric_image}
    I_w^{\mathrm{obj}}(k)
    =M_w(k)\odot I_w(k)
    +\big(1-M_w(k)\big)\odot c,
\end{equation}
where $c\in\mathbb{R}^3$ is a fixed white RGB value used for background replacement and $\odot$ denotes element-wise multiplication.
The same preprocessing operation is used during policy training and deployment.
Figure~\ref{fig:LCP_training} illustrates the two training stages: training the U-Net from generated labels, then training the LCP with the frozen U-Net.

Under ideal foreground extraction, the transformation in \eqref{eq:image_left_action} preserves the task-relevant foreground in the wrist-camera image, while the transformation-dependent background is removed by $\mathcal{P}$.
Therefore, the object-centric observation aims to satisfy the following left-invariance condition:
\begin{equation} \label{eq:object_visual_invariance}
    g_l\circ I_w^{\mathrm{obj}}
    = \mathcal{P}(g_l\circ I_w)
    = \mathcal{P}(I_w)
    = I_w^{\mathrm{obj}}.
\end{equation}
In practice, segmentation errors can leave background content in the image, while illumination and reflections can change the appearance of the retained foreground.
We use photometric augmentation during LCP training to improve robustness to these residual appearance changes as described in Sec.~\ref{sec:lcp_design}.


We use a pretrained DINOv3 vision transformer~\cite{simeoni2025dinov3} for both the raw-image policies in the design study and the object-centric policy in the final system.
While large-scale visual pretraining provides a strong starting representation, the pretrained features may not capture the fine relative pose cues between task objects in the wrist-camera images.
We therefore adapt DINOv3 through layer-wise learning-rate decay (LLRD)~\cite{bao2022beit}, using a decay factor of $0.75$ from the final transformer block toward earlier blocks.
This gives earlier blocks smaller learning rates, allowing the encoder to adapt to the task images while changing those pretrained blocks more conservatively.
Let $\mu_\phi$ denote the resulting adapted visual encoder.
The object-centric wrist image is encoded by $\mu_\phi$ before its features are passed to the LCP action decoder $\mathcal{D}_\psi$.
The front and back wrist-camera images use separate visual encoders with the same architecture.
The backbone and adaptation strategy are evaluated in Sec.~\ref{sec:lcp_design}.

If~\eqref{eq:object_visual_invariance} holds exactly, the visual feature is also left-invariant,
\begin{equation}
    \mu_\phi(g_l\circ I_w^{\mathrm{obj}})
    =\mu_\phi(I_w^{\mathrm{obj}}).
\end{equation}
We do not claim that this equality holds exactly in the real system. Rather, it specifies the visual-representation condition required for the pipeline-level equivariance result in Sec.~\ref{sec:equivariance}; the object-centric preprocessing and adapted pretrained encoder are designed to promote its satisfaction in practice.
Their effects are evaluated in the standardized design study in Sec.~\ref{sec:lcp_design}.

\subsection{Training Procedure}
We first note that Diff-EDF and the LCP are trained separately. We follow the original Diff-EDF's data-collection and training recipes.
The LCP is trained by imitation learning from demonstrations collected at a fixed task configuration.
The final training set is formed through dataset aggregation: after training on the initial demonstrations, additional demonstrations are collected to cover correction states and operating conditions that are insufficiently represented in the initial dataset, and the policy is retrained on the aggregated set.
Note that we pose this procedure as a training practice rather than as a separate algorithmic component of EquiContact.

During training, the task reference frame is known and can be used to calculate the GCEV in~\eqref{eq:GCEV}.
At deployment, however, $\gedf$ is estimated by Diff-EDF and contains estimation error.
To expose the LCP to such errors during training, a bounded $\SE$ perturbation of the task reference frame is sampled once for each demonstration episode and held fixed throughout that episode.
The perturbation is resampled for the episode in each training epoch, and the GCEV is recomputed at every time step from the perturbed reference frame and the recorded end-effector pose.
This preserves temporal consistency of a reference-frame estimation error within an episode while exposing the policy to different reference-frame errors across training.
Further implementation details are provided in Sec.~\ref{sec:lcp_design} and Appendix~\ref{app:LCP_detail}.

\subsection{Main Result: EquiContact Vision-to-Force Equivariance} \label{sec:equivariance}
We now state the vision-to-force equivariance property of EquiContact.
The result relies on two conditions: the $\SE$ equivariance of the reference-frame estimator in~\eqref{eq:equivariance_diff_edf}, and left invariance of the object-centric wrist-camera image $I_w^{\mathrm{obj}}$ in~\eqref{eq:object_visual_invariance}. 
The first condition is provided by the equivariant structure of Diff-EDF, while the second is the image-invariance condition discussed in Sec.~\ref{sec:vision_encoder_left_invariance}.

We first introduce the following lemma.
\begin{lemma}[Left invariance of GCEV $\eg$ and elastic wrench $\fg$~\cite{seo2023contact}]
\label{lemma:1}
For any $g_1,g_2,g_l\in\SE$,
\begin{equation}
    \begin{split}
    \eg(g_lg_1,g_lg_2) &= \eg(g_1,g_2), \\
    \fg(g_lg_1,g_lg_2,K_p,K_R)
    &= \fg(g_1,g_2,K_p,K_R).
    \end{split}
\end{equation}
\end{lemma}
\begin{proof}
    See \cite{seo2023contact}.
\end{proof}


We now introduce the left-invariance property of LCP.

\begin{proposition}[Left-invariance of LCP]
\label{proposition:1}
Suppose the object-centric preprocessing operation $\mathcal{P}$ satisfies \eqref{eq:object_visual_invariance} and left-equivariance of Diff-EDF \eqref{eq:equivariance_diff_edf} is satisfied. Then, the LCP $\pi_\theta$ and its configuration-level representation $\bar{\pi}_\theta$ is left-invariant, i.e.,
\begin{equation} 
    \begin{split}
    \pi_\theta(o) &= \pi_\theta(g_l \circ o), \quad \forall g_l \in \SE.
    \end{split}
\end{equation}    
\end{proposition}
\begin{proof}
    First, the left-transformed observation $g_l \circ o$ reads
    \begin{equation}
        g_l \circ o = (g_l \circ \eg(g,\gedf), g_l \circ F_e, g_l \circ I_{w}),
    \end{equation}
    where $\gedf$ can be written 
    \begin{equation}
        \gedf = \EDF(\calO^{\mathrm{obj}}, \calO^{\mathrm{grasp}}, \calO^{\mathrm{others}}).
    \end{equation}
    Under the left group action, using lemma~\ref{lemma:1} and left-equivariance of Diff-EDF \eqref{eq:equivariance_diff_edf}, $g_l \circ \eg$ now evaluates as 
    \begin{equation} \label{eq:left_invariance_GCEV}
        \begin{split}
            g_l \circ \eg(g,\gedf) &= \eg(g_lg, \EDF(g_l \circ \calO^{\mathrm{obj}}, \calO^{\mathrm{grasp}}, \calO^{\mathrm{others}})) \\
            &= \eg(g_lg, g_l \cdot \EDF(\calO^{\mathrm{obj}}, \calO^{\mathrm{grasp}}, \calO^{\mathrm{others}})) \\
            &= \eg(g_lg, g_l \gedf) = \eg(g,\gedf)
        \end{split}
    \end{equation}
    Under the left group action $g_l$ on observation $o$, together with the assumption of object-centric preprocessing operation $P$ \eqref{eq:object_visual_invariance} and \eqref{eq:left_invariance_GCEV}, it follows that:
    \begin{equation}
        \begin{split}
            \pi_\theta (g_l \circ o) &= \mathcal{D}_\psi(g_l \circ \eg, g_l \circ F_e, \mu_\phi(\mathcal{P}(g_l \circ I_w))) \\
            &= \mathcal{D}_\psi (\eg, F_e, \mu_\phi(I_w^{\mathrm{obj}})) \\
            &= \mathcal{D}_\psi (\eg, F_e, \mu_\phi(\mathcal{P}(I_w))) = \pi_\theta(o).
        \end{split}
    \end{equation}
    Notice that $g_l \circ F_e = F_e$ as $F_e$ is defined on the end-effector body frame.
\end{proof}

Next, we introduce two corollaries. 
For the following corollary, let $\bar{\pi}_\theta(g,g_{\mathrm{ref}},F_e)$, a configuration-representation of policy $\pi_\theta$, denote the output of $\pi_\theta$ using the sensor observations at the indicated end-effector and task configurations, with the surrounding scene held unchanged.

\begin{corollary}[Equivalence of Spatial Randomization \cite{seo2023contact}] \label{corollary:1}
    Under the conditions of Proposition 1, the following equation holds:
    \begin{equation}
        \bar{\pi}_\theta(g,g_l g_{\mathrm{ref}}, F_e) = \bar{\pi}_\theta(g_l^{-1}g, g_{\mathrm{ref}}, F_e).
    \end{equation}
\end{corollary}
\begin{proof}
    From  Proposition~\ref{proposition:1}, the following left-invariance holds for configuration representation $\bar{\pi}_\theta$:
    \begin{equation}
        \bar{\pi}_\theta(g_l \circ (g, g_{\mathrm{ref}},F_e)) = \bar{\pi}_\theta(g, g_{\mathrm{ref}}, F_e).
    \end{equation}
    The proof follows by applying Proposition~\ref{proposition:1} to the configuration $(g,g_l g_{\mathrm{ref}})$ using the common left transformation $g_l^{-1}$. 
\end{proof}
Corollary~\ref{corollary:1} generalizes Proposition~1 of~\cite{seo2023contact} and establishes the equivalence between task reference-frame randomization and end-effector pose randomization.
Under the left-invariance condition, demonstrations can be collected at a fixed task configuration, while initial end-effector pose randomization provides coverage of different relative configurations. From the practical perspective, this simplifies data collection, since teleoperation at tilted task configurations is more challenging.

The second Corollary states the equivariance of LCP, if it is represented using the spatial frame.

\begin{corollary}[Left Equivariance of LCP] \label{corollary:2}
Consider the LCP represented in the spatial frame $\pi_\theta^s$, defined as
\begin{equation}
    \pi_\theta^s(g, o) \triangleq (g g_{\mathrm{rel}}, K_p, K_R),
\end{equation}
where $(g_{\mathrm{rel}}, K_p, K_R) = \pi_\theta(o)$.
Then, $\pi_\theta^s$ satisfies the following equivariance property:
\begin{equation}
    \begin{split}
        \pi^s_\theta(g_l \circ (g, o)) &= g_l \circ \pi^s_\theta(g,o) \\
        &= (g_l g_d, K_p, K_R)
    \end{split}
\end{equation}
\end{corollary}
\begin{proof}
    Noticing that the group action $g_l$ acting on $g$ transforms $g$ into $g_lg$, and together with the left-invariance property from Proposition~\ref{proposition:1}, it is straightforward to show that
    \begin{equation}
        \begin{split}
            \pi_\theta^s(g_l g, g_l\circ o) &= \left((g_lg)\cdot g_{\mathrm{rel}}, K_p, K_R \right) \\
            &= \left(g_l g_d, K_p, K_R\right).
        \end{split}
    \end{equation}
    Therefore, when the policy is left-transformed by an arbitrary element $g_l \in \SE$, the resulting desired pose commands in the spatial frame $g_d$ are also transformed to $g_l g_d$, showing the equivariance property.
\end{proof}

Building on these results, we finally state the vision-to-force equivariance property of EquiContact.
Let $h_\Theta$ denote the body-frame elastic-wrench map induced by the complete EquiContact pipeline, with output $\fg$ as defined in \eqref{eq:GIC_wrench}.
The main proposition is stated below.
\begin{proposition} \label{proposition:2}
    Suppose the conditions of Proposition~\ref{proposition:1} are satisfied. 
    Then, the body-frame elastic wrench map induced by EquiContact is left-invariant, and the corresponding elastic wrench expressed in the spatial frame is $\SE$ equivariant.
\end{proposition}
\begin{proof}
    Let body-frame elastic wrench map $h_\Theta$ be defined as
    \begin{equation}
        h_\Theta(g,o) \triangleq \fg (g,g_d,K_p,K_R), \quad (g_d, K_p, K_R) = \pi_\theta^s(g,o)
    \end{equation}
    Then, under the left group action $g_l$, together with the result of Lemma~\ref{lemma:1} and Corollary~\ref{corollary:2}, the following equality holds:
    \begin{equation}
        \begin{split}
            h_\Theta( g_l &\circ(g,o)) = h_\Theta(g_l g , g_l \circ o)\\
            &= \fg(g_lg, g_lg_d, K_p, K_R) \\
            &= \fg(g,g_d, K_p, K_R) = h_\Theta(g, o)
        \end{split}
    \end{equation}
    which proves the invariance property of the body-frame wrench map. Finally, as the $h_\Theta$ is left-invariant and is defined on the end-effector frame, its spatial frame representation $h_\Theta^s$ is given by \cite{seo2023contact}
    \begin{equation}
        h_\Theta^s(g_l \circ(g(k),o(k))) = \Ad_{g_l^{\text{-}1}}^\top h_\Theta^s(g(k),o(k)),
    \end{equation}
    where $\Ad$ is a large Adjoint map \cite{seo2023geometric}. By the definition of equivariance \eqref{eq:left_equivariance} and noticing that $\Ad_{g_l^{\text{-}1}}^\top$ is a representation of $g_l$ acting on the wrench, the wrench map of EquiContact expressed in the spatial frame $h_\Theta^s$ is equivariant.
\end{proof}

\section{Empirical Realization of the Localized Compliant Policy}\label{sec:lcp_design}
In this section, we present  results of a study designed to test several LCP components, particularly those involving the visual backbone.
We first standardize the evaluation protocol, select a pretrained visual backbone, compare adaptation strategies, and investigate the effect of object-centric preprocessing. The results are summarized in Table~\ref{table:lcp_design}.

\subsection{Standardized Design-Study Protocol}
The geometric structure of the LCP specifies which quantities should remain unchanged under a common left transformation, but it does not determine the visual backbone, the form of task adaptation, or the action-generation model. We therefore realize the learned components of the LCP through a sequential design study in which each stage fixes the choice selected in the preceding stage.

Unless otherwise stated, all models in this study use the circular peg-in-hole dataset with $149$ demonstrations and the same training recipe. 
During demonstration collection and robot evaluation, the initial end-effector pose is randomized within $\pm 8\,\mathrm{cm}$ in translation and $\pm 15^\circ$ in rotation relative to the nominal initial pose.

Separately, during LCP training, the task reference frame is perturbed within $\pm 5\,\mathrm{cm}$ in translation and $\pm 15^\circ$ in rotation per axis.
The reference-frame perturbation is sampled once per episode and held fixed throughout the trajectory.
Models are trained for $74$ epochs ($174$k optimizer steps) using a warm-up followed by the $\cos4$ learning-rate schedule, photometric augmentation, and gradient clipping \cite{dasari2024ingredients}. The policy peak learning rate is $4\times10^{-5}$, while trained vision-encoder parameters use $1\times10^{-5}$. We use ACT as the common action-generation model for the visual study. After encoder adaptation, it performs reliably in the nominal condition but still exhibits failures under large task transformations, allowing the effects of the design choices involving the vision encoder backbone to be highlighted.

Robot evaluations are organized into progressively more challenging task conditions: Flat GT verifies basic task execution at the nominal platform orientation, i.e., the trained case (Fig.~\ref{fig:equi_contact}, top-right); Tilt $45^\circ$ (Fig.~\ref{fig:extreme_transformations}, right), GT tests spatial and visual generalization under a large task transformation while removing reference-frame estimation error; and Tilt $45^\circ$, Diff-EDF additionally tests recovery from the imperfect task frame encountered in the deployed pipeline. The design study is sequential: each stage is evaluated on the conditions needed to answer its designated design question, and later stages retain the preceding conditions while adding the next evaluation condition. Therefore, we state that entries marked ``NT'' (not tested) in Table~\ref{table:lcp_design} are intentionally not tested at that stage rather than missing results.

The standardized dataset is intentionally collected with substantial background variation and task-irrelevant visual distractors (Fig.~\ref{fig:scene_randomization}) so that the raw-image baselines are not trained in a visually static scene. During development, demonstrations are iteratively aggregated until reliable, near-perfect execution is obtained in the nominal flat condition; the resulting $149$-demonstration dataset is then fixed for the standardized study. Together with photometric augmentation, this provides a deliberately strong raw-image baseline before introducing any explicit object-centric preprocessing.

\begin{table*}[t]
    \setlength\doublerulesep{0.5pt}
    \renewcommand\tabularxcolumn[1]{m{#1}}
    \footnotesize
    \centering
    \vspace{-10pt}
    \caption{
    Sequential LCP design study. The evaluation conditions are introduced progressively: Flat GT tests nominal task execution, Tilt $45^\circ$ GT tests generalization under a large task transformation without reference-frame estimation error, and Tilt $45^\circ$ Diff-EDF additionally tests robustness to the estimated task frame used at deployment. NT (Not Tested) indicates conditions not evaluated at the corresponding design stage. Raw uses the original wrist image and Painted uses U-Net object-centric preprocessing. Stages 1--3 use ACT; Stage 4 uses Flow Matching. We report success rate (SR), in (successful trials / total trials).}
    \label{table:lcp_design}
    \vspace{-4pt}
    \begin{tabularx}{\linewidth}{
     >{\centering\arraybackslash\hsize=1.20\hsize}X
     >{\raggedright\arraybackslash\hsize=1.40\hsize}X
     >{\centering\arraybackslash\hsize=0.60\hsize}X
     >{\centering\arraybackslash\hsize=0.80\hsize}X
     >{\centering\arraybackslash\hsize=0.95\hsize}X
     >{\centering\arraybackslash\hsize=1.05\hsize}X
    }
    \toprule[1pt]\midrule[0.3pt]
    \addlinespace[3pt]
    Stage & {Backbone / adaptation} & Image & {Flat, GT} & {Tilt $45^\circ$, GT} & {Tilt $45^\circ$, Diff-EDF} \\
    \addlinespace[1pt]
    \midrule \midrule
    \multirow{3}{*}{1: backbone}
    & CLIP-RN50, frozen & Raw & 0 / 10 & NT & NT \\
    & SigLIP2, frozen & Raw & 2 / 10 & NT & NT \\
    & DINOv3, frozen & Raw & \textbf{4 / 10} & NT & NT \\
    \midrule
    \addlinespace[2pt]
    \multirow{3}{*}{2: adaptation}
    & DINOv3, last block & Raw & 10 / 10 & 6 / 10 & NT \\
    & DINOv3, LLRD & Raw & 10 / 10 & \textbf{8 / 10} & NT \\
    & DINOv3, LoRA-r32 & Raw & 10 / 10 & 6 / 10 & NT \\
    \midrule
    \addlinespace[2pt]
    \multirow{3}{*}{3: object-centric}
    & DINOv3, last block & Painted & 10 / 10 & \textbf{10 / 10} & 8 / 10 \\
    & DINOv3, LLRD & Painted & 10 / 10 & 9 / 10 & \textbf{10 / 10} \\
    & DINOv3, LoRA-r32 & Painted & 10 / 10 & 9 / 10 & 8 / 10 \\
    \midrule
    \addlinespace[2pt]
    4: Flow-Matching
    & \textbf{DINOv3, LLRD} & \textbf{Painted} & \textbf{10 / 10} & \textbf{10 / 10} & \textbf{10 / 10} \\
    \midrule[0.3pt]\bottomrule[1pt]
    \end{tabularx}
\end{table*}

\subsection{Selecting the Pretrained Visual Backbone}
We first compare frozen CLIP-RN50, SigLIP2, and DINOv3 encoders using raw wrist images. Among these encoders, DINOv3 achieves the highest success rate on the nominal flat task, with $4/10$ successful trials, compared with $2/10$ for SigLIP2 and $0/10$ for CLIP-RN50. Qualitatively, the frozen DINOv3 policy generally approaches the target but fails to achieve the final alignment accuracy required for insertion. We therefore select DINOv3 as the pretrained backbone for the remaining studies.

\subsection{Adapting DINOv3 for Precise Manipulation}
The poor absolute performance of the frozen encoders indicates that large-scale pretraining alone is not sufficient to handle domain and task gap between internet-scale large field-of-view images and close-range wrist-camera observations.
We therefore compare three adaptation strategies for DINOv3: fine-tuning only the final transformer block, layer-wise learning-rate decay (LLRD)~\cite{bao2022beit}, and LoRA applied across all $12$ transformer blocks~\cite{hu2022lora} with rank $32$ (LoRA-r32).

Adaptation is the largest single improvement in the standardized study. On the flat robot test, all three adapted models achieve $10/10$, whereas frozen DINOv3 achieves $4/10$. Under the $45^\circ$ outward tilt with a ground-truth reference frame, the raw-image LLRD model achieves $8/10$, while last-block fine-tuning and LoRA-r32 each achieve $6/10$. These results show that task-specific adaptation substantially improves insertion reliability, while the choice of adaptation method has a smaller effect in these tests. The raw-image models also demonstrate some robustness under large task transformations, even without explicit object-centric preprocessing.

\subsection{Object-Centric Visual Invariance}
The preceding adaptation study improves task performance, but the raw-image models still depend on transformation-dependent background content. We therefore apply the same U-Net background painting to each adapted DINOv3 encoder. On the $45^\circ$ ground-truth-frame test, the pooled success rate increases from $20/30$ with raw images to $28/30$ with painted images. This improvement is observed despite the background-diverse data collection and photometric augmentation already used by the raw-image baselines, indicating that explicitly removing task-irrelevant background provides additional robustness under large task transformations.

We next evaluate the painted models using the same ten Diff-EDF estimates of the $45^\circ$ tilted hole. This condition is particularly relevant to EquiContact because the LCP is trained to refine an imperfect reference frame rather than to receive the exact task frame. LLRD succeeds in all $10$ trials, compared with $8/10$ for last-block fine-tuning and $8/10$ for LoRA-r32. The two most difficult estimates have the largest rotational errors, and one additionally contains a $57.5\,\mathrm{mm}$ local translational error, outside the $\pm 5\,\mathrm{cm}$ training range. LLRD is the only model that recovers from both estimates.

Because the adaptation methods show mixed success rates under ground-truth and Diff-EDF reference frames, we also compare their contact-force profiles during insertion.
Table~\ref{table:lcp_force} summarizes the contact profiles for the same Diff-EDF trials. 
LLRD completes the trials in fewer steps and produces lower lateral force and torque than the other adapted models, indicating better alignment in the earlier stage of the insertion.
We therefore select LLRD for the final LCP encoder.

\begin{table}[t]
    \setlength\doublerulesep{0.5pt}
    \renewcommand\tabularxcolumn[1]{m{#1}}
    \footnotesize
    \centering
    \caption{
    Contact behavior of painted DINOv3 models on the same ten $45^\circ$ Diff-EDF estimates. Peak values are the median of the per-trial peaks. Lower force, torque, and trial length indicate cleaner recovery from reference-frame error.}
    \label{table:lcp_force}
    \vspace{-4pt}
    \begin{tabularx}{\columnwidth}{
     >{\raggedright\arraybackslash\hsize=1.10\hsize}X
     >{\centering\arraybackslash\hsize=0.70\hsize}X
     >{\centering\arraybackslash\hsize=0.70\hsize}X
     >{\centering\arraybackslash\hsize=1.50\hsize}X
     >{\centering\arraybackslash\hsize=1.00\hsize}X
    }
    \toprule[1pt]\midrule[0.3pt]
    \addlinespace[3pt]
    Adaptation & SR & Steps & {$\|F_{xy}\|$ mean / peak} & {$\|\tau\|$ mean} \\
    \addlinespace[1pt]
    \midrule \midrule
    Last block & 8 / 10 & 468 & 2.7 / 9.1 N & 1.71 Nm \\
    \midrule
    \addlinespace[2pt]
    LoRA-r32 & 8 / 10 & 406 & 2.5 / 7.8 N & 1.53 Nm \\
    \midrule
    \addlinespace[2pt]
    \textbf{LLRD} & \textbf{10 / 10} & \textbf{355} & \textbf{1.1 / 5.4 N} & \textbf{1.04 Nm} \\
    \midrule[0.3pt]\bottomrule[1pt]
    \end{tabularx}
\end{table}

\subsection{Action Generation Model Selection}
We next evaluate ACT and conditional Flow Matching using the selected object-centric DINOv3 encoder adapted with LLRD. Both models use the LCP observations and predict the relative pose and stiffness commands defined in~\eqref{eq:LCP}. Therefore, the choice of action-generation model does not change the geometric interface used in the equivariance analysis.

With the selected visual representation, the ACT-based LCP succeeds in $10/10$ flat GT, $9/10$ Tilt $45^\circ$ GT, and $10/10$ Tilt $45^\circ$ Diff-EDF trials.
The final Flow-Matching LCP succeeds in $10/10$ trials under all three conditions. 
We therefore select the Flow-Matching as the action-generation model of LCP, and use it in the subsequent system evaluations.

\section{System Evaluation}\label{sec:experimental_result}
We evaluate EquiContact through real-robot experiments addressing three research questions:
\begin{itemize}[leftmargin=25pt,topsep=0pt]
\setlength{\itemsep}{0pt}
\item[\textbf{RQ1}] Is an equivariant high-level reference-frame estimator followed by geometric admittance control sufficient for tight insertion, or is learned local refinement necessary?
\item[\textbf{RQ2}] What is the role of compliance in achieving reliable contact-rich manipulation?
\item[\textbf{RQ3}] Can anchoring a localized compliant policy on an equivariantly estimated reference frame enable spatial generalization to out-of-distribution task configurations without retraining?
\end{itemize}

All experiments are conducted on a real robot. In the full-pipeline evaluations, Diff-EDF supplies estimated task reference frames to the LCP. Evaluations intended to isolate the LCP use ground-truth (GT) task reference frames. For each trial, the initial end-effector pose is randomized within $\pm 8\,\mathrm{cm}$ in translation and $\pm 15^\circ$ in rotation relative to its nominal initial pose, consistent with the demonstration-collection protocol.
Training and implementation details are provided in Appendix~\ref{app:implementation_details_extended}.

\subsection{Peg-in-Hole Benchmark Results}

\subsubsection{Necessity of Learned Local Refinement (RQ1)}
We first examine whether an equivariant global reference-frame estimate alone is sufficient for tight insertion. Diff-EDF estimates the insertion frame $\gedf$, which is directly tracked through GAC with a learned gain-scheduling policy~\cite{seo2023contact}, without the LCP. This baseline fails on all \textbf{0 / 10} flat-platform trials. The estimated frame does not consistently achieve the millimeter-level accuracy required for direct insertion; see Appendix~\ref{app:diff_edf_error}. This motivates a learned policy that can refine the coarse reference using local visual and force feedback.

\subsubsection{Role of Compliance (RQ2)}
\begin{figure}[t]
    \centering
    \includegraphics[width=\columnwidth]{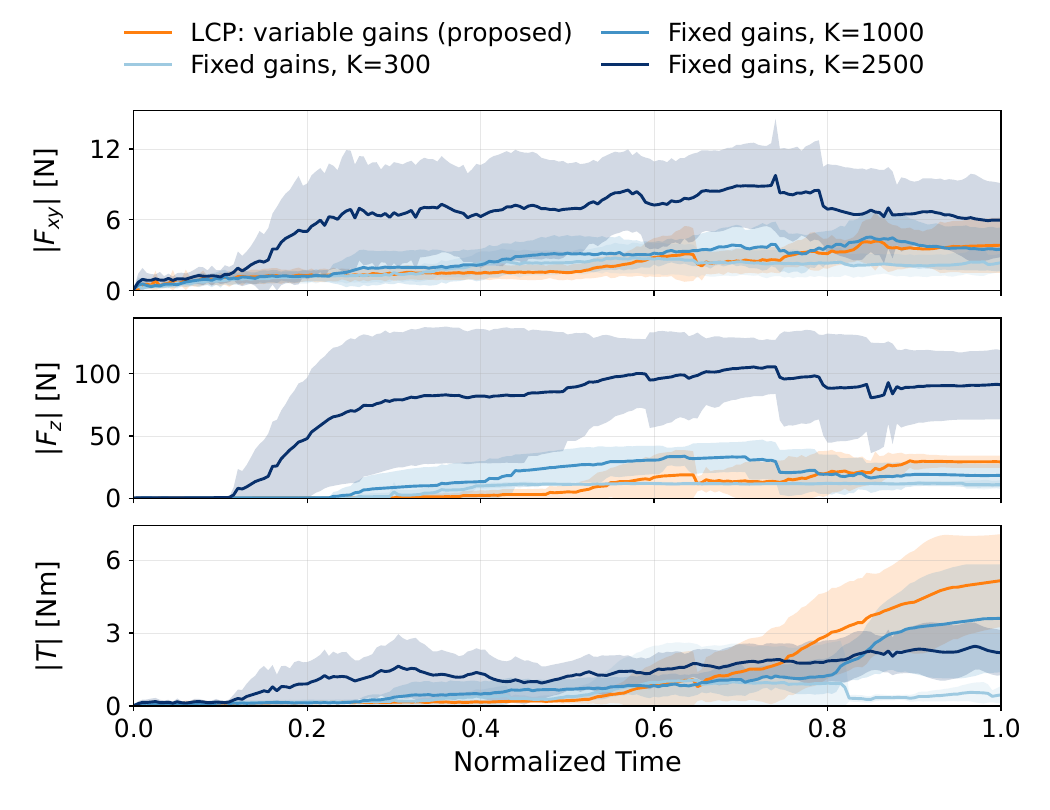}
    \vspace{-25pt}
    \caption{Normalized-time force--torque profiles for triple-peg insertion with variable and fixed gains.}
    \label{fig:flat_gains_ft_normtime}
\end{figure}

We evaluate the role of compliance on the orientation-sensitive triple-peg insertion task. The LCP predicts variable gains within $300$--$1500$. For the fixed-gain baselines, we set $K_p=K_R=KI_3$, with $K\in\{300,1000,2500\}$. In particular, the high-gain setting ($K=2500$) is intended to approximate non-compliant control by enforcing stiff end-effector motion. Table~\ref{table:compliance_ablation} reports insertion success and mean absolute axial force ($|F_z|$) and Fig.~\ref{fig:flat_gains_ft_normtime} shows the force/torque profiles during insertion.

\begin{table}[t]
    \setlength\doublerulesep{0.5pt}
    \renewcommand\tabularxcolumn[1]{m{#1}}
    \setlength{\tabcolsep}{3pt}
    \footnotesize
    \centering
    \caption{Compliance ablation on flat triple-peg insertion. The variable LCP uses gains of $300$--$1500$; mean and standard deviation (Std) of the absolute axial force $|F_z|$ are reported.}
    \label{table:compliance_ablation}
    \vspace{-4pt}
    \begin{tabularx}{\columnwidth}{
     >{\raggedright\arraybackslash\hsize=1.15\hsize}X
     >{\centering\arraybackslash\hsize=0.55\hsize}X
     >{\centering\arraybackslash\hsize=1.30\hsize}X
    }
    \toprule[1pt]\midrule[0.3pt]
    \addlinespace[2pt]
    {Gain setting} & {SR} & {Mean $\pm$ Std $|F_z|$ [N]} \\
    \addlinespace[1pt]
    \midrule \midrule
    \textbf{Variable (ours)} & \textbf{10 / 10} & $\mathbf{18.6 \pm 7.6}$ \\
    \midrule
    Fixed $K=300$ & 0 / 10 & $11.3 \pm 1.0$ \\
    Fixed $K=1000$ & 8 / 10 & $24.8 \pm 8.0$ \\
    Fixed $K=2500$ & 4 / 10 & $93.4 \pm 27.4$ \\
    \midrule[0.3pt]\bottomrule[1pt]
    \end{tabularx}
\end{table}

The low fixed gain produces the smallest mean axial force ($11.3\,\mathrm{N}$) but fails all $10$ trials; during execution, we observed that it could not maintain the alignment needed for insertion, although it maintains low interaction force.
Fixed $K=1000$ succeeds in \textbf{8 / 10} trials, whereas $K=2500$ succeeds in only \textbf{4 / 10} and increases mean axial force to $93.4\,\mathrm{N}$, approximately five times the LCP value of $18.6\,\mathrm{N}$. The variable-gain LCP achieves \textbf{10 / 10} success without the excessive axial forces observed at high fixed gain. These results show that compliant execution needs to balance sufficient stiffness for insertion with tolerance to residual pose error.

\subsubsection{Spatial Generalization of EquiContact (RQ3)}
To answer RQ3, we first compare the global policy and the LCP using ground-truth task reference frames (Table~\ref{table:representation_ablation}). The global policy receives the current end-effector pose $g$ and task reference frame $g_{\mathrm{ref}}$, both expressed in the spatial frame, the F/T measurement $F_e$, and the wrist-camera images $I_w$. It predicts a desired end-effector pose $g_d$ in the spatial frame and admittance gains $K_p$ and $K_R$.
\begin{table}[t]
    \setlength\doublerulesep{0.5pt}
    \renewcommand\tabularxcolumn[1]{m{#1}}
    \setlength{\tabcolsep}{2.5pt}
    \footnotesize
    \centering
    \caption{Effect of geometric policy representation on spatial generalization. Both policies use wrist camera images, force/torque feedback, predicted admittance gains, and GAC. Nominal denotes the untransformed flat configuration; Flat OOD denotes translated flat configurations. Results report successful trials / total trials.}
    \label{table:representation_ablation}
    \vspace{-4pt}
    \begin{tabularx}{\columnwidth}{
     >{\raggedright\arraybackslash\hsize=1.10\hsize}X
     >{\raggedright\arraybackslash\hsize=1.34\hsize}X
     >{\raggedright\arraybackslash\hsize=0.90\hsize}X
     >{\centering\arraybackslash\hsize=0.83\hsize}X
     >{\centering\arraybackslash\hsize=0.83\hsize}X
    }
    \toprule[1pt]\midrule[0.3pt]
    \addlinespace[3pt]
    {Method} & {Observation} & {Action} & {Flat (nominal)} & {Flat OOD} \\
    \addlinespace[1pt]
    \midrule \midrule
    Global policy + GAC
    & Global pose, task reference, FT
    & Global pose
    & 10 / 10
    & 0 / 10 \\
    \midrule
    \addlinespace[2pt]
    \textbf{LCP + GAC (ours)}
    & \textbf{GCEV, FT}
    & \textbf{Relative pose}
    & \textbf{10 / 10}
    & \textbf{20 / 20} \\
    \midrule[0.3pt]\bottomrule[1pt]
    \end{tabularx}
\end{table}

Both policies succeed in \textbf{10 / 10} trials at the nominal flat configuration. Under translated flat OOD configurations, the global policy succeeds in \textbf{0 / 10} trials, whereas the LCP succeeds in \textbf{20 / 20}.
The global policy faces two sources of distribution shift under task translation. First, its geometric inputs $g$ and $g_{\mathrm{ref}}$ are expressed in the spatial frame, so the transformed tasks produces pose inputs not seen in the demonstrations collected on the nominal task configuration.
Second, its desired pose $g_d$ is also expressed in the spatial frame and normalized using statistics from those demonstrations. Therefore, actions required at the transformed configurations can fall outside the action range seen during training. The LCP instead uses GCEV and relative pose actions, which avoid these spatial-frame shifts in the geometric policy interface.


We next evaluate the complete pick-and-insertion EquiContact pipeline at transformed task configurations. Diff-EDF estimates reference frames for both grasping and insertion, and the LCP performs the local execution at each stage. These trials test two things: 1. whether EquiContact can complete the full task and 2. whether the LCP can operate from estimated reference frames rather than ground-truth frames. The latter requires the LCP to correct task reference frame estimation error during execution. Table~\ref{table:success_rate_full_pipeline} reports the circular and triple-peg results.
\begin{table}[!t]
    \setlength\doublerulesep{0.5pt}
    \renewcommand\tabularxcolumn[1]{m{#1}}
    \footnotesize
    \centering
    \caption{Full-pipeline EquiContact success rates with Diff-EDF in the loop. Circular PiH uses the original Diff-EDF candidate post-processing; triple PiH uses the subsequent consensus post-processing. Results are reported separately.}
    \label{table:success_rate_full_pipeline}
    \vspace{-4pt}
    \begin{tabularx}{\columnwidth}{
     >{\raggedright\arraybackslash\hsize=1.40\hsize}X
     >{\centering\arraybackslash\hsize=0.80\hsize}X
     >{\centering\arraybackslash\hsize=0.80\hsize}X
    }
    \toprule[1pt]\midrule[0.3pt]
    \addlinespace[3pt]
    {Test scenario} & {Circular PiH, SR} & {Triple PiH, SR} \\
    \addlinespace[1pt]
    \midrule \midrule
    Flat (OOD) & 20 / 20 & 20 / 20 \\
    \addlinespace[2pt]
    Tilted ($30^\circ$, OOD) & 10 / 10 & 10 / 10 \\
    \addlinespace[2pt]
    Tilted ($45^\circ$, OOD) & 9 / 10 & 10 / 10 \\
    \midrule[0.3pt]\bottomrule[1pt]
    \end{tabularx}
\end{table}



For circular peg-in-hole (PiH), EquiContact succeeds in \textbf{20 / 20} translated-flat trials, \textbf{10 / 10} trials at $30^\circ$, and \textbf{9 / 10} at $45^\circ$. In the single failed trial, the Diff-EDF insertion-frame estimate has an approximately $45^\circ$ yaw error, beyond the $\pm15^\circ$ rotational perturbations used to train the LCP.
We also evaluate triple-peg insertion, where all three pegs must align with their corresponding holes at the same time. This makes insertion sensitive to errors in every rotational direction. EquiContact succeeds in \textbf{20 / 20} translated-flat trials, \textbf{10 / 10} at $30^\circ$, and \textbf{10 / 10} at $45^\circ$.

The full-pipeline results show that EquiContact can complete pick-and-insertion at transformed task configurations, with the LCP operating on Diff-EDF reference-frame estimates for both grasping and insertion. The failed circular-peg trial also shows a limit of this correction: its approximately $45^\circ$ yaw error falls outside the rotational perturbations used to train the LCP.

\subsection{Additional Tasks and Stress Tests}

We further evaluate the LCP in two settings beyond the full-pipeline pick-and-insertion tasks. First, large triple-peg tilts test its execution as the wrist-camera view changes substantially with task orientation. Second, surface wiping tests whether the same LCP design can be trained for a different contact-rich task using new task-specific demonstrations. Both evaluations use ground-truth task reference frames to remove Diff-EDF estimation error from these tests. Table~\ref{table:additional_tasks} summarizes the results.

\begin{table}[t]
    \setlength\doublerulesep{0.5pt}
    \renewcommand\tabularxcolumn[1]{m{#1}}
    \footnotesize
    \centering
    \caption{Additional task and extreme-transformation results for the LCP. All conditions use ground-truth task reference frames to isolate low-level policy generalization from Diff-EDF. The $45^\circ$ triple-peg full-pipeline result is reported in Table~\ref{table:success_rate_full_pipeline}.}
    \label{table:additional_tasks}
    \vspace{-4pt}
    \begin{tabularx}{\columnwidth}{
     >{\raggedright\arraybackslash\hsize=0.85\hsize}X
     >{\raggedright\arraybackslash\hsize=1.35\hsize}X
     >{\centering\arraybackslash\hsize=0.80\hsize}X
    }
    \toprule[1pt]\midrule[0.3pt]
    \addlinespace[3pt]
    {Task} & {Test condition} & {SR} \\
    \addlinespace[1pt]
    \midrule \midrule
    Triple PiH & $-x$ tilt, $90^\circ$ & 10 / 10 \\
    \addlinespace[2pt]
    Triple PiH & $-x$ tilt, $135^\circ$ & 10 / 10 \\
    \addlinespace[2pt]
    Triple PiH & $+y$ tilt, $45^\circ$ & 10 / 10 \\
    \midrule
    \addlinespace[2pt]
    Wiping & Flat (OOD) & 10 / 10 \\
    \addlinespace[2pt]
    Wiping & Tilted ($30^\circ$, OOD) & 10 / 10 \\
    \addlinespace[2pt]
    Wiping & Tilted ($45^\circ$, OOD) & 10 / 10 \\
    \midrule[0.3pt]\bottomrule[1pt]
    \end{tabularx}
\end{table}

\begin{figure}[t]
    \centering
    \includegraphics[width=\columnwidth]{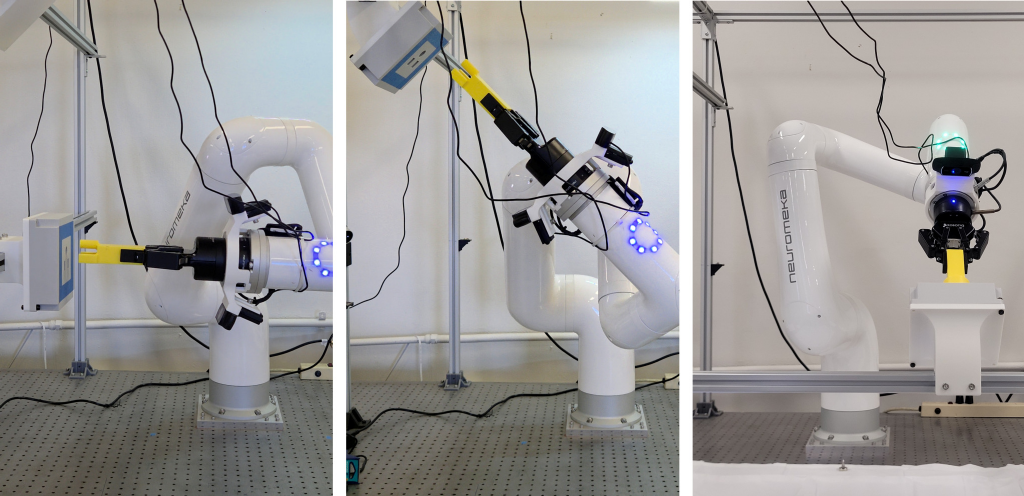}
    \caption{Representative extreme task transformations used to evaluate the LCP. Left: $-x$ axis, $90^\circ$ tilt. Middle: $-x$ axis $135^\circ$ tilt. Right: $+y$ axis, $45\circ$ tilt.}
    \label{fig:extreme_transformations}
\end{figure}

For the orientation-sensitive triple-peg task, the LCP succeeds in \textbf{10 / 10} trials at each of $90^\circ$ and $135^\circ$ (Fig.~\ref{fig:extreme_transformations}). The $45^\circ$ Diff-EDF condition, already reported in Table~\ref{table:success_rate_full_pipeline}, is not repeated in Table~\ref{table:additional_tasks}. Under GT reference frame, the GCEV, body-frame F/T sensor feedback, and relative pose actions preserve their geometric consistency even as wrist-camera observations change substantially. These stress tests demonstrate transfer of the LCP under large task transformations, not the operating range of Diff-EDF.

\begin{figure}[t]
    \centering
    \includegraphics[width=\columnwidth]{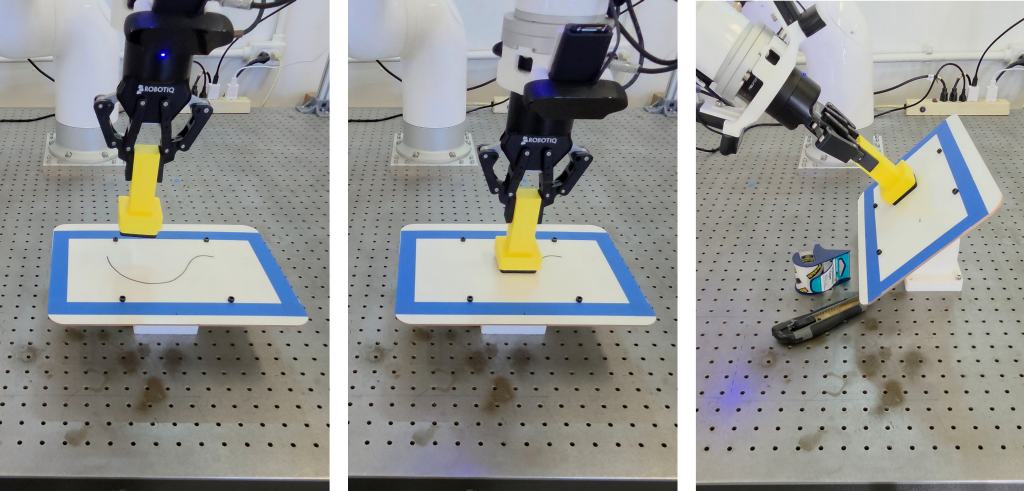}
    \caption{Surface-wiping task. \textbf{Left and Center:} Erasing marker lines on the board. \textbf{Right:} Wiping with the board tilted by $45^\circ$.}
    \label{fig:additional_tasks}
\end{figure}

For surface wiping (Fig.~\ref{fig:additional_tasks}), demonstrations are collected at a fixed flat configuration, and the LCP is evaluated on translated-flat, $30^\circ$-tilted, and $45^\circ$-tilted OOD configurations. The policy initially showed sensitivity to viewpoint-dependent glare on the white board at $30^\circ$. Adding glare variation to the photometric augmentation, without collecting tilted demonstrations or changing the reference-frame perturbation scheme, resolves this failure mode in the reported tests. The resulting LCP succeeds in \textbf{10 / 10} trials in each configuration (\textbf{30 / 30} overall). This illustrates the importance of visual appearance coverage even when geometric observations and actions are structured to preserve transformation properties.

\subsection{Limitations}
\textbf{Controlled Environment.}
Although EquiContact achieves high success rates on contact-rich tasks with spatial generalization, the experiments remain partially controlled. Salient task features, including blue markings on some platforms and distinct object colors, are used to support reliable local visual feedback. Evaluation in less engineered scenes with more natural object appearance remains an important direction.

\textbf{Approximate Visual Invariance.}
The geometric components of EquiContact, including GCEV, body-frame force--torque feedback, relative actions, and GAC, are explicitly structured for equivariance. In contrast, invariance of the wrist-image representation is only approximate. Object-centric preprocessing removes a major source of transformation-dependent background variation, but does not guarantee invariance to foreground appearance changes caused by viewpoint, illumination, shadows, or camera exposure.

\textbf{Bounded Reference-Frame Correction.}
The LCP is trained to refine a coarse task-frame estimate over a bounded error distribution rather than to recover from arbitrary reference-frame errors. The single failure in the $45^\circ$ circular-peg evaluation occurs when Diff-EDF produces approximately $45^\circ$ yaw error, far outside the broad $\pm15^\circ$ rotational perturbation used in LCP training. Although the circular mating geometry is yaw-symmetric, this symmetry is not explicitly encoded by the fixed task-frame convention, GCEV distribution, or camera-specific visual encoders.

\textbf{Task-dependent Procedures}
Introducing a new task requires generating segmentation masks for its task-relevant objects to train the lightweight U-Net. In addition, each LCP in this study is trained and evaluated on a single task; we have not evaluated a single LCP across multiple tasks.

\section{Conclusion}\label{sec:conclusion}
In this work, we presented EquiContact, a hierarchical $\SE$ vision-to-force equivariant framework for spatially generalizable contact-rich manipulation.
EquiContact combines an equivariant high-level reference-frame estimator with a localized compliant policy (LCP) that handles fine contact-rich execution through end-effector-frame geometric feedback and relative pose actions.
We show that this hierarchical structure induces pipeline-level $\SE$ equivariance under equivariant reference-frame estimation and left-invariant visual features.
To address the visual-invariance condition in practice, we introduce object-centric visual preprocessing and systematically investigate the design of the learned visual representation.


On real-robot peg-in-hole tasks with $1\,\mathrm{mm}$ clearance, the full pipeline succeeds in $39/40$ circular-peg and $40/40$ triple-peg trials across translated-flat, $30^\circ$ tilted, and $45^\circ$ tilted configurations, with the LCP trained at a single task configuration.
With ground-truth reference frames, the LCP succeeds in $20/20$ triple-peg trials at $90^\circ$ and $135^\circ$ tilts and in $30/30$ surface-wiping trials.
These results show that a local policy without built-in equivariance, anchored on an equivariant reference frame and executed through left-invariant compliance, can transfer contact-rich behavior learned at one configuration to unseen task configuration.

\bibliographystyle{IEEEtran}
\bibliography{references}

\appendices
\section{Implementation Details}\label{app:implementation_details_extended}

\subsection{Localized Compliant Policy}\label{app:LCP_detail}
To maintain the final LCP's $30\,\mathrm{Hz}$ update rate, we use CUDA Graphs for Flow-Matching sampling to reduce repeated execution overhead. During training, the final Flow-Matching model uses the broad reference-frame perturbation described in Sec.~\ref{sec:lcp_design} for most episodes. In $10\%$ of episodes, it instead uses a small zero-centered Gaussian perturbation to include near-accurate task frames in training. Gradient clipping is set to $1.0$.

\begin{figure}[t]
    \centering
    \includegraphics[width=0.485\columnwidth]{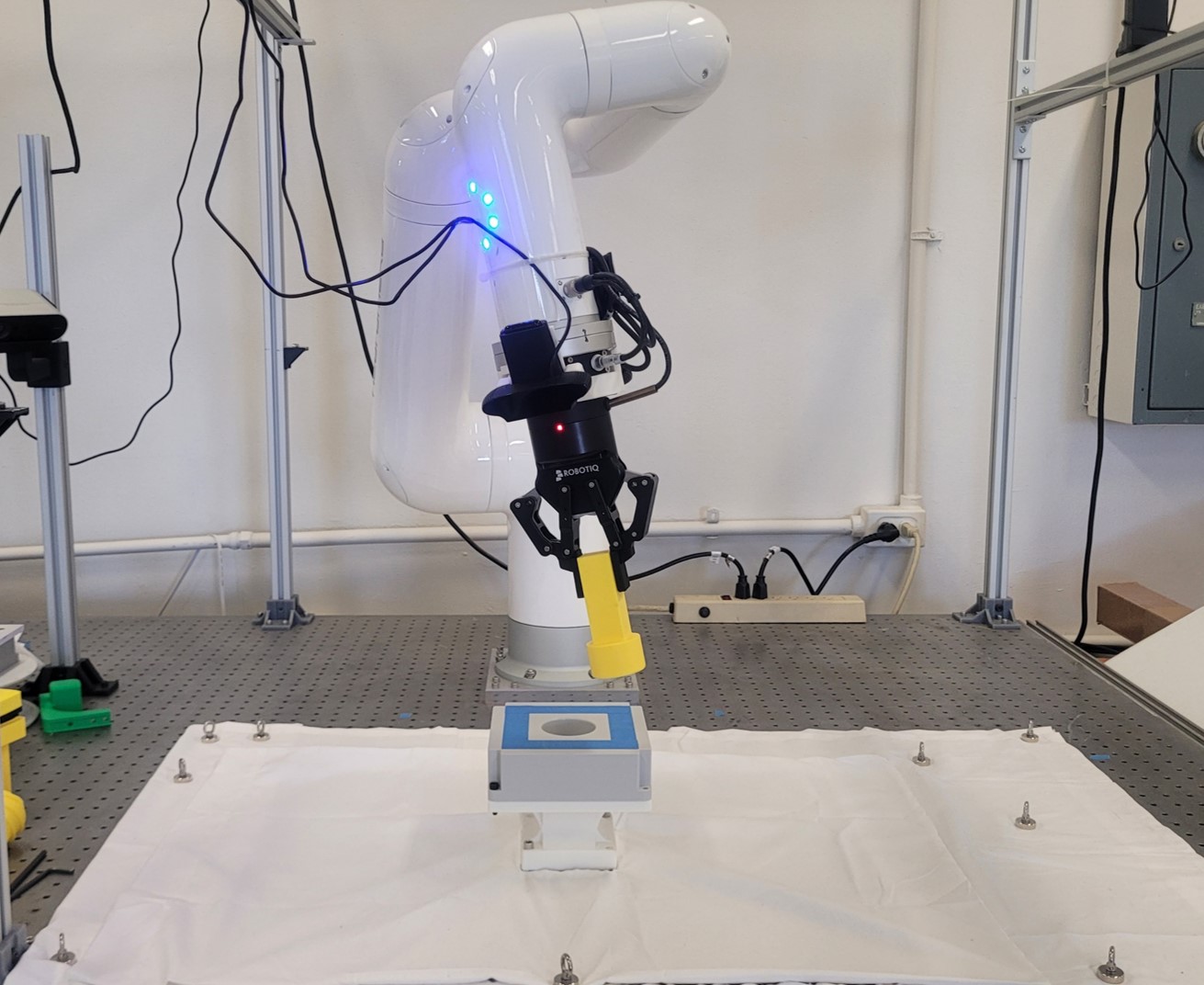}
    \hfill
    \includegraphics[width=0.485\columnwidth]{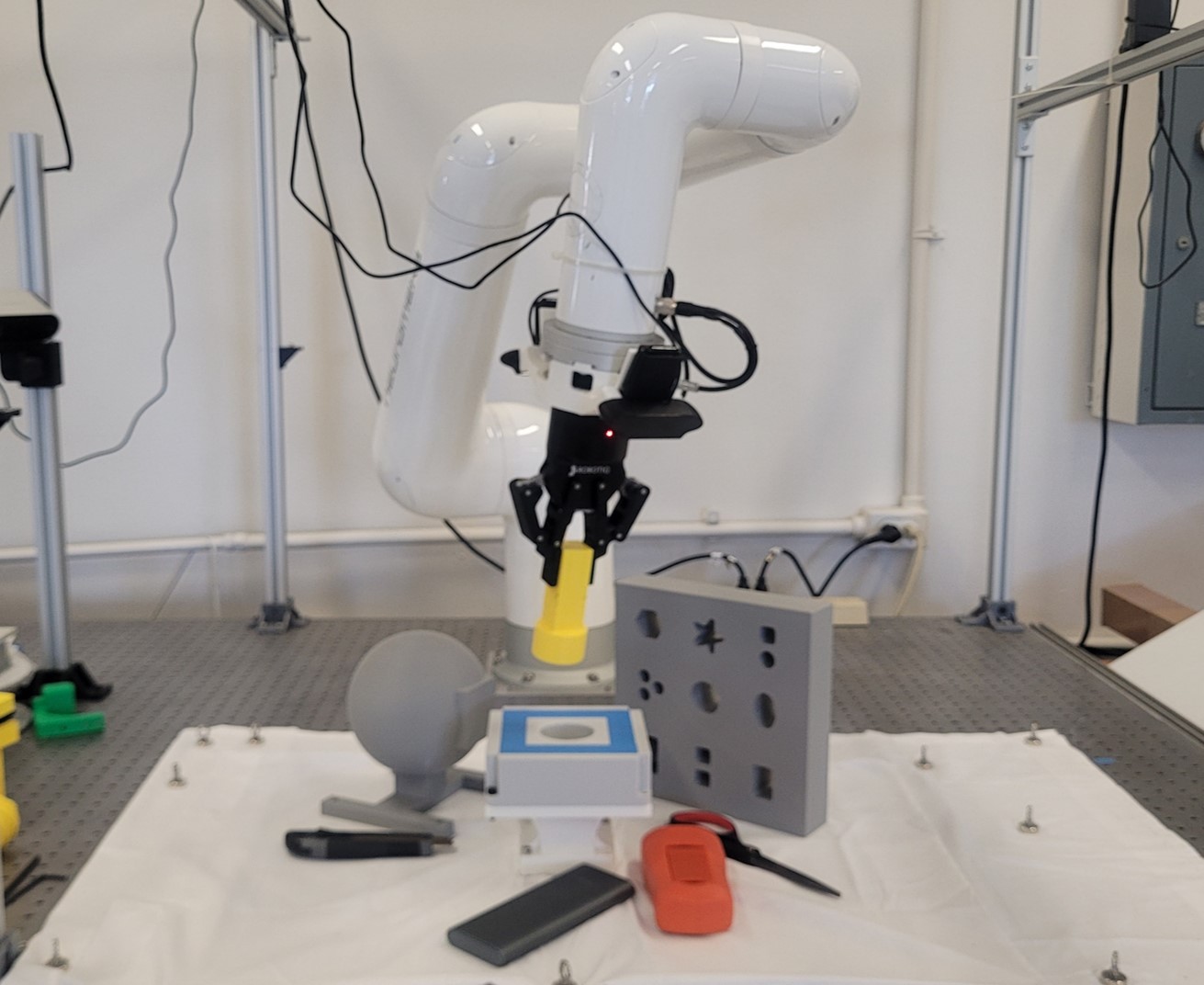}

    \vspace{5pt}

    \includegraphics[width=0.485\columnwidth]{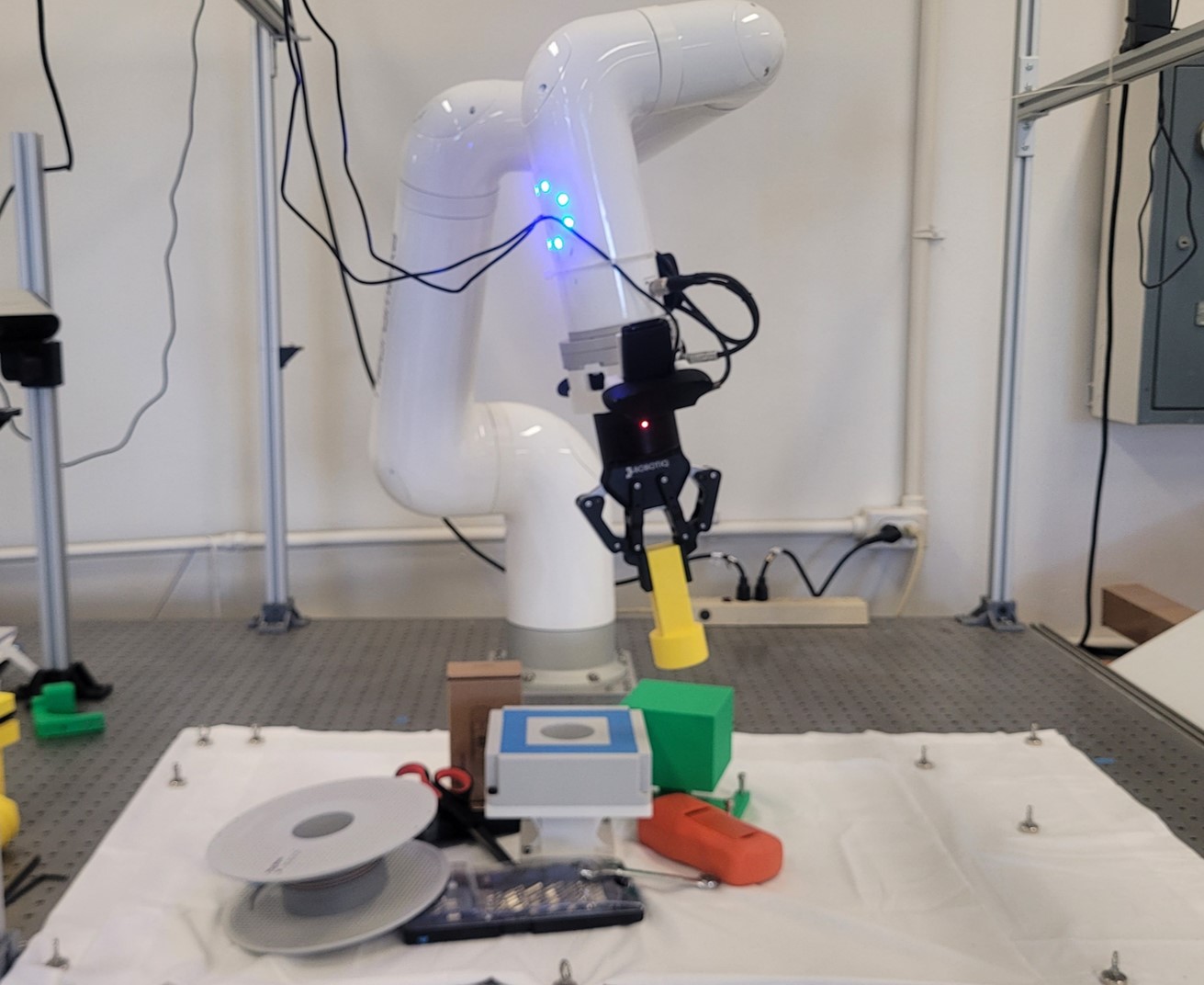}
    \hfill
    \includegraphics[width=0.485\columnwidth]{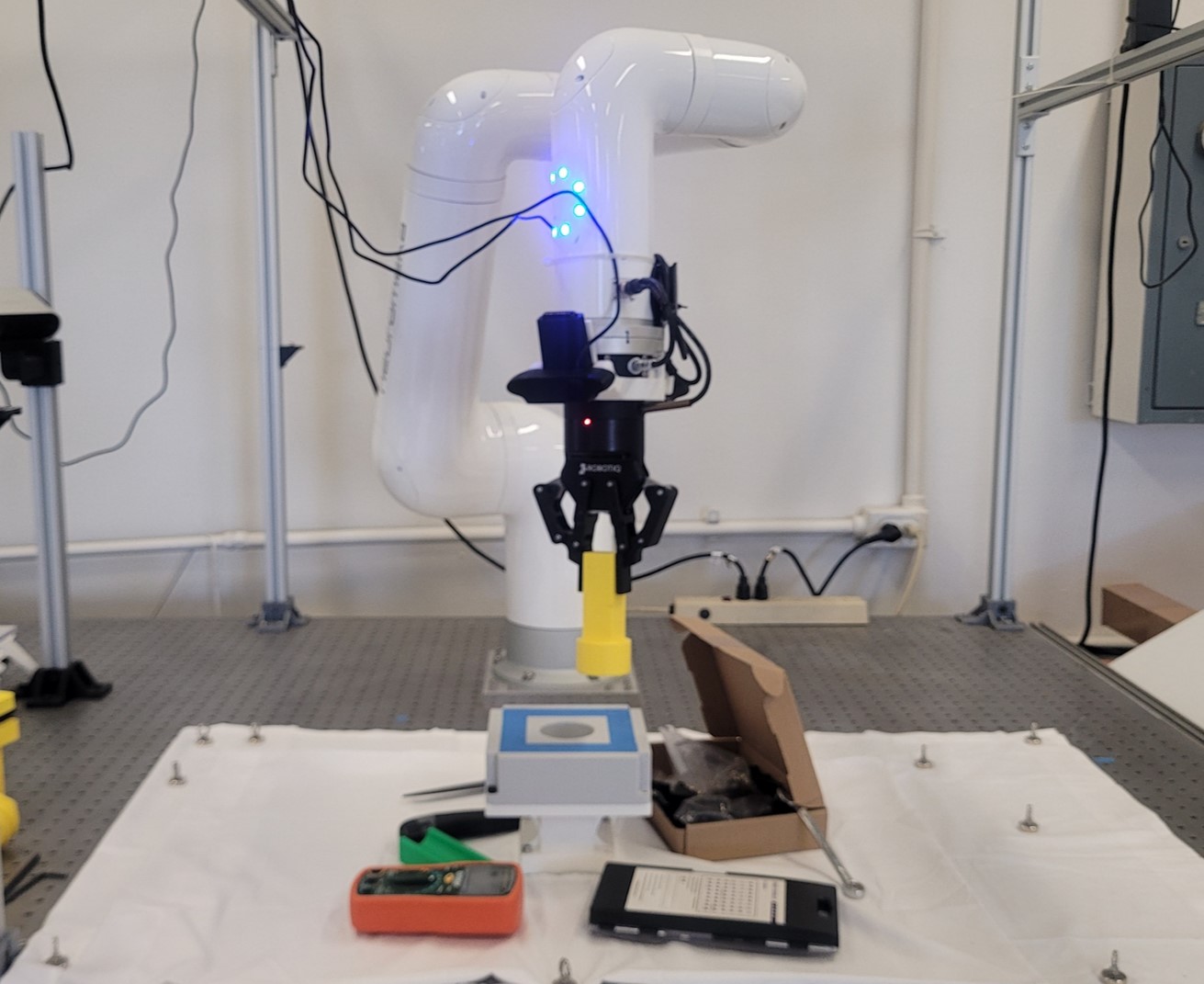}

    \caption{
    Representative scenes from the LCP demonstration dataset.
    Demonstrations are collected at a fixed task configuration with variations in background objects and task-irrelevant visual content. The dataset also includes scenes without the white cotton backgrounds shown in these examples.
    }
    \label{fig:scene_randomization}
\end{figure}

Table~\ref{table:dataset_statistics} summarizes the demonstrations used for the three task-specific LCPs, and Fig.~\ref{fig:scene_randomization} shows representative scenes from the circular peg-in-hole dataset.
\begin{table}[t]
    \setlength\doublerulesep{0.5pt}
    \renewcommand\tabularxcolumn[1]{m{#1}}
    \setlength{\tabcolsep}{2.5pt}
    \footnotesize
    \centering
    \caption{
    Demonstration dataset statistics for the three
    task-specific LCPs. Demonstrations are recorded at
    $30\,\mathrm{Hz}$. The reported statistics exclude
    the additional $60$ terminal-pose hold steps appended
    to each episode during training dataset construction.
    Episode lengths are reported in steps.
    }
    \label{table:dataset_statistics}
    \vspace{-4pt}
    \begin{tabularx}{\columnwidth}{
      >{\raggedright\arraybackslash\hsize=1.15\hsize}X
      >{\centering\arraybackslash\hsize=0.65\hsize}X
      >{\centering\arraybackslash\hsize=0.90\hsize}X
      >{\centering\arraybackslash\hsize=1.30\hsize}X
      >{\centering\arraybackslash\hsize=1.00\hsize}X
    }
    \toprule[1pt]\midrule[0.3pt]
    \addlinespace[3pt]
    Task & Episodes & Frames &
    Mean $\pm$ SD & $[\min, \max]$ \\
    \addlinespace[1pt]
    \midrule \midrule
    Circular PiH & 149 & 66,443 &
        $445.9 \pm 61.5$ & $[313,671]$ \\
    \addlinespace[3pt]
    Triple PiH & 98 & 52,878 &
        $539.6 \pm 125.9$ & $[354,1116]$ \\
    \addlinespace[3pt]
    Wiping & 105 & 68,819 &
        $655.4 \pm 166.6$ & $[250,998]$ \\
    \midrule[0.3pt]\bottomrule[1pt]
    \end{tabularx}
\end{table}

\textbf{Admittance Gains and Demonstration Collection.} 
During demonstration collection, the teleoperator commands end-effector motion with a SpaceMouse and switches among four preset gain modes. For both learning and GAC execution, only the diagonal entries of $K_p$ and $K_R$ are used, so the stiffness command is represented by six values. The preset modes are
\begin{itemize}[leftmargin=*]
    \item Low-gain: $(300,300,300,300,300,300)$,
    \item High-gain: $(1000,1000,1000,1000,1000,1000)$,
    \item Contact: $(1500,1500,300,1500,1500,1500)$,
    \item Insertion: $(300,300,1500,300,300,300)$.
\end{itemize}
The first three entries correspond to translational stiffness and the last three to rotational stiffness. The desired inertia is set to $M=0.5I_{6\times6}$. These values are implementation dependent and were tuned on the real robot.

For peg insertion, high gain is used for free-space alignment, contact mode for surface contact and local search, and insertion mode once the peg is partially inserted. The same interface is used to collect wiping demonstrations, with gain-mode transitions chosen by the teleoperator according to the contact phase.

\subsection{Object-Centric Segmentation}\label{app:unet_training}
For each task, we train a four-class U-Net with $1.94$M parameters. Recordings are split by episode, holding out every tenth episode for segmentation evaluation. Each original time step provides two wrist-camera images. The U-Net is trained for $8$ epochs using pixel-wise cross-entropy loss, AdamW (learning rate $10^{-3}$, weight decay $10^{-4}$), batch size $32$, and a one-cycle learning-rate schedule. We deploy the final-epoch checkpoint and held-out episodes are not used for checkpoint selection. This split applies only to segmentation: LCP training uses all task demonstrations, with no policy validation split and the final policy checkpoint deployed.

For circular peg-in-hole, the split contains $134$ training episodes ($119{,}304$ images). The triple-peg segmentation model is trained jointly on circular- and triple-peg recordings, totaling $222$ training episodes ($213{,}496$ images). The wiping segmentation model is trained on $94$ demonstration episodes together with
$9$ segments of one additional recording in which the wiper is held away from the board,
totaling $103$ training episodes ($129{,}556$ images). This additional recording is used
only for segmentation training.

\begin{figure}[t]
    \centering
    \includegraphics[width=\columnwidth]{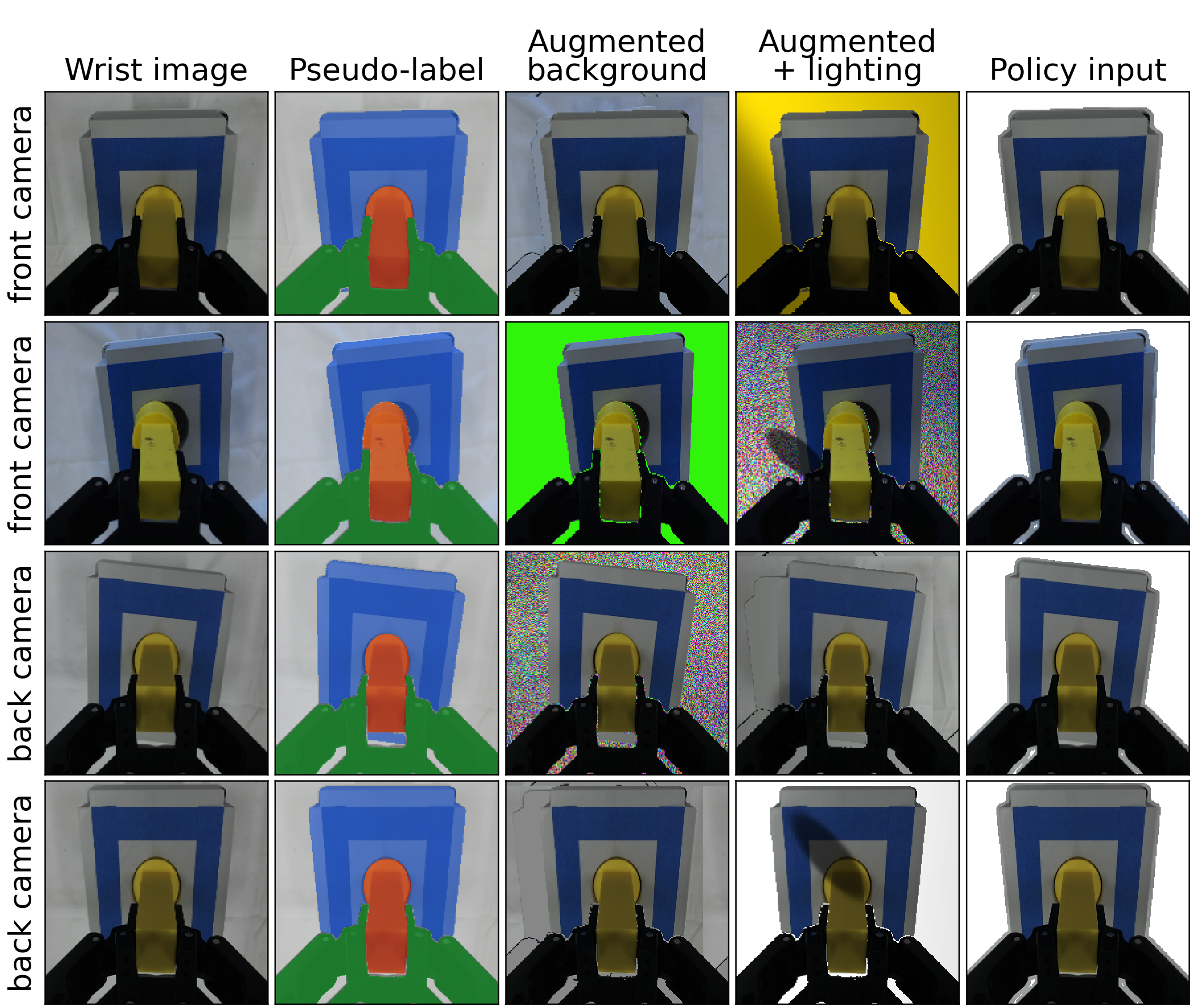}
    \caption{Examples of object-centric preprocessing and augmentation for peg-in-hole wrist images. From left to right: original wrist image, generated segmentation labels, background replacement, additional lighting variation, and the resulting policy input.}
    \label{fig:unet_background_aug}
\end{figure}

For the peg-in-hole tasks, pseudo-labels are generated offline using Grounded-DINO and SAM2. For wiping, labels are generated using color- and template-based rules. Frames without usable labels are excluded from U-Net training. Background replacement and lighting augmentation are applied during U-Net training, as illustrated in Fig.~\ref{fig:unet_background_aug}. The wiping LCP additionally uses glare augmentation on the board.

\subsection{Diff-EDF Implementation}\label{app:diff_edf_implementation}
Diff-EDF is trained from scene and grasp point clouds paired with target reference frames. We collect $20$ Diff-EDF demonstrations: $10$ with the flat platform and $10$ with the tilted platform, with randomized target translations and rotations. The remaining training procedure follows~\cite{diff_edf}.

For this task setup, separate pick-style Diff-EDF models are used to estimate the grasping and insertion reference frames. The peg is initially upright, and the grasping policy is trained to produce a consistent peg/end-effector alignment. We therefore do not explicitly model the right-equivariance term associated with arbitrary relative transformations between the grasped object and end-effector.

At inference, Diff-EDF produces multiple candidate poses. We observed that direct aggregation can be sensitive to isolated orientation outliers. We therefore apply consensus post-processing across the candidate poses before aggregation, suppressing candidates that are inconsistent with the dominant pose cluster. The final reference frame is reconstructed from the consensus candidates; for grasping, the known upright-orientation assumption is retained.

\subsection{GAC Implementation}\label{app:GAC_implementation}
GAC realizes~\eqref{eq:desired_dynamics} by generating pose commands for the robot's lowest-level pose-tracking controller. In discrete time,
\begin{align}\label{eq:gac_implementation}
    V_d^b(k)
    &=V^b(k)+T_sM^{-1}
    \left(F_e(k)-\fg(k)-K_dV^b(k)\right), \nonumber\\
    \tilde g_d(k)
    &=g(k)\exp\!\left((V_d^b(k))^\wedge \cdot T_s\right),
\end{align}
where $T_s=5\,\mathrm{ms}$ and $(\cdot)^\wedge:\mathbb{R}^6\rightarrow\mathfrak{se}(3)$ denotes the twist hat map. Thus, GAC is executed at $200\,\mathrm{Hz}$, independently of the approximately $30\,\mathrm{Hz}$ LCP policy update rate.

\section{Additional Experimental Results}

\subsection{Diff-EDF Reference-Frame Error}\label{app:diff_edf_error}
\begin{table}[h!]
    \setlength\doublerulesep{0.5pt}
    \renewcommand\tabularxcolumn[1]{m{#1}}
    \centering
    \caption{Training-set RMSE of Diff-EDF reference-frame estimates. The dimensions of translational errors in $x,y,z$ directions, given by $e_{T,x}, e_{T,y}, e_{T,z}$, are $\mathrm{mm}$ and the rotational errors in $x,y,z$ directions, given by $e_{R,x}, e_{R,y}, e_{R,z}$, are $\mathrm{deg}$.}
    \label{table:RMSE_diff_edf}
    \begin{tabularx}{\linewidth}{
     >{\centering\arraybackslash\hsize=1.0\hsize}X >{\centering\arraybackslash \hsize=1.0\hsize}X >{\centering\arraybackslash \hsize=1.0\hsize}X >
     {\centering\arraybackslash \hsize=1.0\hsize}X >
     {\centering\arraybackslash \hsize=1.0\hsize}X >
     {\centering\arraybackslash \hsize=1.0\hsize}X >
     {\centering\arraybackslash \hsize=1.0\hsize}X
    }
    \toprule[1pt]\midrule[0.3pt]
      & $e_{T,x}$ & $e_{T,y}$ & $e_{T,z}$ & $e_{R,x}$ & $e_{R,y}$ & $e_{R,z}$ \\
    \midrule
    pick & 7.173 & 6.933 & 6.199 & 7.650 & 15.90 & 15.67 \\
    place & 13.75 & 8.241 & 5.999 & 3.806 & 5.560 & 5.660 \\
    \midrule[0.3pt]\bottomrule[1pt]
    \end{tabularx}
\end{table}
Table~\ref{table:RMSE_diff_edf} reports the training-set RMSE of the Diff-EDF reference-frame estimates used in the original system characterization.

The translational error is substantially larger than the approximately $1\,\mathrm{mm}$ clearance scale of the circular peg-in-hole task, explaining why Diff-EDF followed only by compliant pose tracking is insufficient for reliable insertion. The LCP is therefore trained to refine the coarse estimate using local visual and force feedback.

\end{document}